\documentclass[journal]{IEEEtran}
\usepackage[utf8]{inputenc}
\usepackage{amsmath}
\usepackage{times}
\usepackage{graphicx}
\usepackage{color}
\usepackage{multirow}
\usepackage[ruled]{algorithm}
\usepackage[noend]{algpseudocode}
\usepackage{pifont}
\usepackage{tabu}
\usepackage{array,multirow, makecell}
\usepackage{subcaption}
\usepackage{latexsym}
\usepackage{color,soul}
\usepackage{mathtools}
\usepackage{wrapfig}
\usepackage{amssymb}

\usepackage{hyperref}
\usepackage[utf8]{inputenc}
\usepackage[T1]{fontenc} 
\usepackage{amsmath}
\usepackage{amssymb}
\usepackage{tabu}
\usepackage{array,multirow, makecell}
\usepackage{latexsym}
\usepackage{color,soul}
\usepackage{mathtools}
\usepackage{wrapfig}
\usepackage{footmisc}

\ifCLASSINFOpdf
\else
\fi
\begin{document}
%
\title{Towards Scalable, Efficient and Accurate Deep Spiking Neural Networks with Backward Residual Connections, Stochastic Softmax and Hybridization}
%
%
%

\author{Priyadarshini~Panda, \textit{Member}, \textit{IEEE}, Aparna Aketi,
        and Kaushik~Roy,~\IEEEmembership{Fellow,~IEEE}
\thanks{P. Panda is with the Electrical Engineering Department at Yale University, New Haven, CT, 06510. A. Aketi, and K. Roy are with the School
of Electrical and Computer Engineering, Purdue University, West Lafayette,
IN, 47907 USA. This work was done while P. Panda was a PhD student at Purdue University. E-mail: priya.panda@yale.edu}
}

%
%

\markboth{Journal of \LaTeX\ Class Files,~Vol.~11, No.~4, December~2012}%
{Shell \MakeLowercase{\textit{et al.}}: Bare Demo of IEEEtran.cls for Journals}
%



\maketitle

\begin{abstract}
Spiking Neural Networks (SNNs) may offer an energy-efficient alternative for implementing deep learning applications. In recent years, there have been several proposals focused on supervised (conversion, spike-based gradient descent) and unsupervised (spike timing dependent plasticity) training methods to improve the accuracy of SNNs on large-scale tasks. However, each of these methods suffer from \textit{scalability, latency and accuracy} limitations. In this paper, we propose novel algorithmic techniques of modifying the SNN configuration with \textit{backward residual connections, stochastic softmax and hybrid artificial-and-spiking neuronal activations} to improve the learning ability of the training methodologies to yield competitive accuracy, while, yielding large efficiency gains over their artificial counterparts. Note, artificial counterparts refer to conventional deep learning/artificial neural networks. Our techniques apply to VGG/Residual architectures, and are compatible with all forms of training methodologies. Our analysis reveals that the proposed solutions yield near state-of-the-art accuracy with significant energy-efficiency and reduced parameter overhead translating to hardware improvements on complex visual recognition tasks, such as, CIFAR10, Imagenet datatsets.
\end{abstract}


%
\IEEEpeerreviewmaketitle

\section{Introduction}

Neuromorphic computing, specifically, Spiking Neural Networks (SNNs) have become very popular as an energy-efficient alternative for implementing standard artificial intelligence tasks \cite{indiveri2011frontiers, pfeiffer2018deep, cao2015spiking, panda2016unsupervised, sengupta2016hybrid}. Spikes or binary events drive communication and computation in SNNs that not only is close to biological neuronal processing, but also offer the benefit of event-driven hardware operation \cite{ankit2017resparc, indiveri2015neuromorphic}. This makes them attractive for real-time applications where power consumption and memory bandwidth are important factors. What is lacking, however, is proper training algorithms that can make SNNs perform at par with conventional artificial neural networks (ANNs). Today, there is a plethora of work detailing different algorithms or learning rules for implementing deep convolutional spiking architectures for complex visual recognition tasks \cite{panda2016unsupervised,diehl2015fast, lee2016training, o2013real, kheradpisheh2018stdp, masquelier2009competitive, lee2018training, lee2018deep, srinivasan2018stdp, panda2017asp, diehl2015unsupervised, masquelier2007unsupervised,hunsberger2015spiking, bellec2018long,neftci2019surrogate, mostafa2017supervised, sengupta2019going, severa2019training, srinivasan2019restocnet}. Most algorithmic proposals focus on integrating the discrete or discontinuous spiking behavior of a neuron in a supervised or unsupervised learning rule. All proposals maintain overall sparse network activity (implies low power operation) while improving the accuracy (implies better performance) on image recognition applications (mostly, benchmarked against state-of-the-art datasets like Imagenet \cite{deng2009imagenet}, CIFAR \cite{krizhevsky2010convolutional}, MNIST \cite{lecun2010mnist}).

Collating the previous works, we can broadly categorize the SNN training methodologies into three types: 1) Conversion from artificial-to-spiking models \cite{sengupta2019going, diehl2015fast}, 2) Approximate Gradient Descent (AGD) based backpropagation with spikes (or accounting temporal events) \cite{lee2016training,neftci2019surrogate}, and 3) Unsupervised Spike Timing Dependent Plasticity (STDP) based learning \cite{diehl2015unsupervised, srinivasan2018stdp}. Each technique presents some advantages and some disadvantages. While conversion methodology has yielded state-of-the-art accuracies for large datasets like Imagenet on complex architectures (like VGG \cite{simonyan2014very}, ResNet \cite{he2016deep}), the latency incurred to process the rate-coded image\footnote{SNNs process event data obtained with rate or temporal coding instead of real-valued pixel data. Rate coding is widely used for SNN applications, where, a real-valued pixel data is converted to a Poisson-distribution based spike train with the spiking frequency proportional to the pixel value \cite{diehl2015unsupervised}. That is, high valued pixels output more spikes and vice-versa.\label{foot1}} is very high \cite{pfeiffer2018deep, sengupta2019going, lee2019enabling}. AGD training addresses the latency concerns yielding $\sim10-15\times$ benefits as compared to the conversion \cite{lee2019enabling, bellec2018long, neftci2019surrogate}. However, AGD still lags behind conversion in terms of accuracy for larger and complex tasks. 
The unsupervised STDP training, while being attractive for real-time hardware implementation on several emerging and non-von Neumann architectures \cite{ankit2017resparc, sengupta2017encoding, wang2017memristors,van2017non, perez2010neuromorphic, linares2011spike}, also suffers from accuracy/scalability deficiencies.
 
From the above discussion, we can gather that addressing \textit{Scalability, Latency and Accuracy} issues are key towards achieving successful SNN methodologies. In this paper, we precisely address each of these issues through the lens of \textit{network architecture modification}, \textit{softmax classifier adaptation} and \textit{network hybridization with a mix of Rectified Linear Unit/ReLU (or ANN-like) and Leaky-Integrate-and-Fire (or SNN-like) neuronal activations in different layers}. 

\section{Related Work, Motivation and Contributions}

\subsection{Addressing Scalability with Backward Residual Connections}
Scalability limitations of STDP/AGD approaches arises from their depth incompatibility with deep convolutional networks which are necessary for achieving competitive accuracies. SNNs forward propagate spiking information and thus require sufficient spike activity across all layers of a deep network to conduct training. However, previous works have shown that spiking activity decreases drastically for deeper layers of a network (that we define as \textit{vanishing spike propagation}), thereby, causing training issues for networks with large number of layers \cite{kheradpisheh2018stdp, masquelier2009competitive, lee2016training, lee2018deep, srinivasan2018stdp,panda2016unsupervised,diehl2015fast}. 

\begin{figure}
\centering
\includegraphics[width=\linewidth]{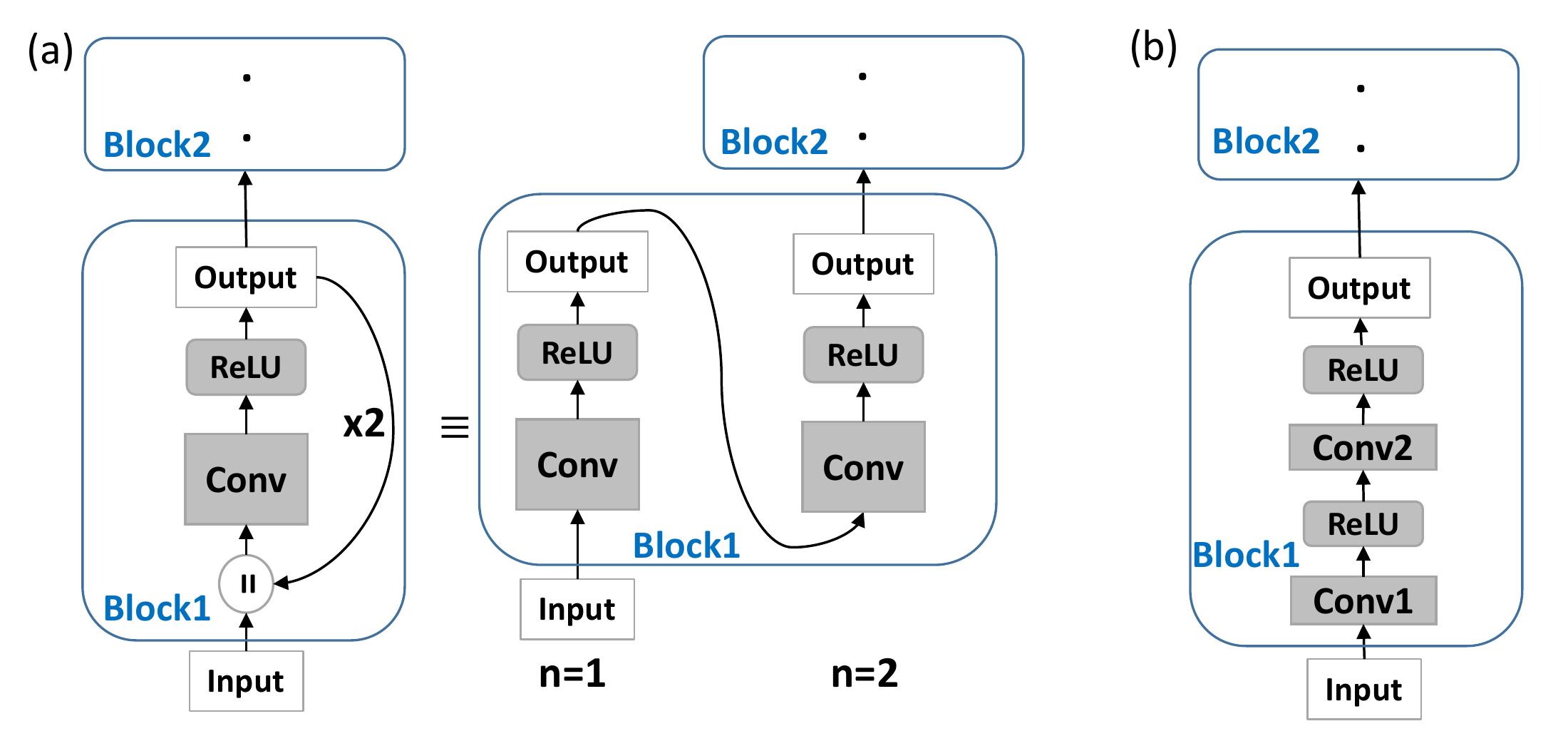}
\caption{(a) A simple neural network architecture with Backward Residual (BackRes with $n=2$) connections is shown. The layers in $Block1$ are unrolled $n=2$ times to perform the BackRes computation with weights of $Conv$ layer reused at each unrolling step. (b) A network with 2 unique convolutional layers in $Block1$ is shown. Note, BackRes computation in (a) achieves the same logical depth of the network in (b).}
\end{figure}

From ANN literature, it is known that depth is key to achieving improved accuracy for image recognition applications \cite{lecun2015deep, szegedy2015going}. Then, the question arises, \textit{can we modify the spiking network architecture to be less deep without compromising accuracy}? Kubilius et al. \cite{kubilius2018cornet} proposed Core Object Recognition or CORnet models (with what we term as \textit{backward residual connections}) that transform deep feedforward ANN models into shallow recurrent models. Fig. 1 illustrates the Backward Residual (BackRes) block architecture. It is similar to that of a recurrent network unrolled over time with weights shared over repeated computations of the output. Specifically, the computations in $Block1$ are performed twice before processing $Block2$. For $n=1$, $Block1$ processes original input information, while, for $n=2$, the same $Block1$ with repeated weights processes the output from previous step. Note, the original input is processed only once for $n=1$. For $n>1$, the block processes its output from the previous step. Essentially, BackRes connections enable a network to achieve similar logical depth as that of a deep feedforward network without introducing additional layers. The 1-convolutional layer block in Fig. 1 (a) achieves the logical depth of a 2-convolutional layer block as shown in Fig. 1(b) and is expected to achieve near iso-accuracy with that of the 2-convolutional layer block$\footnote{There is a limit to which BackRes compensates for depth diversity with iso-accuracy. VGG2x8 network with 2 convolutional layers unrolled 8 times may suffer accuracy loss as compared to a VGG16 network with 16 convolutional layers. But, VGG2x4 may yield near iso-accuracy as VGG8. Note, VGG2x4 and VGG8 have same logical depth of 8 convolutional layers.}$. 
The BackRes connection brings two key advantages: 1) Reduction in the total number of parameters since we are reusing the same weights over multiple steps of unrolling, 2) Diversification of gradient update for each unrolled step due to different input-output combinations. 

\textit{Our Contribution:} We utilize BackRes connections and the diversified gradients to enable training of logically deep SNN models with AGD or STDP that otherwise cannot be trained (with multiple layers) due to vanishing spike propagation. Further, we show that converting a deep ANN (with BackRes blocks) into a deep SNN necessitates the use of multiple threshold-spiking neurons per BackRes block to achieve lossless conversion. We also demonstrate that BackRes SNN models (say, VGG2x4) yield both lower memory complexity (proportional to number of weights/parameters) and sparser network activity with decreased computational overhead (proportional to total inference energy) as compared to a deep architecture (say, VGG8) of similar logical depth across different SNN training methodologies.  

\subsection{Addressing Latency with \underline{Stoch}astic Soft\underline{max} (Stochmax)}
In order to incur minimal loss during pixel-to-spike conversion with rate coding\footref{foot1} (generally, used in all SNN experiments), the number of time steps of the spike train has to sufficiently large. This, in turn, increases the latency of computation. Decreasing the latency implies larger loss in image-to-spike conversion that can result in lower accuracy. 

Across all SNN training methodologies, the final classifier or output layer which yields the prediction result is usually a softmax layer similar to that of an ANN. It is general practice, in SNN implementation, to collect all the accumulated spiking activity over a given time duration from the penultimate layer of a deep SNN and feed it to a softmax layer that calculates the loss and prediction based on the integrated spike information \cite{lee2019enabling,masquelier2007unsupervised, lee2016training}. While the softmax classifier based training has produced competitive results, the latency incurred still is significantly high. The question that arises here is, \textit{‘Can we compensate for reduced latency (or, higher loss during image-to-spike conversion) by improving the learning capability of the SNN by augmenting the softmax functionality?’} Lee et al. \cite{lee2018dropmax} proposed a stochastic version of a softmax function (\textit{stochmax}) that drops irrelevant (non-target) classes with adaptive dropout probabilities to obtain improved accuracy in ANN implementations. Stochmax can be viewed as a stochastic attention mechanism, where, the classification process at each training iteration selects a subset of classes that the network has to attend to for discriminating against other false classes. For instance, while training for a \textit{cat} instance, it is useful to train the model with more focus on discriminating against confusing classes, such as, \textit{jaguar, tiger} instead of orthogonal classes like \textit{truck, whale}. Softmax, on the other hand, collectively optimizes the model for target class (\textit{cat}) against all remaining classes (\textit{jaguar, tiger, truck, whale}) in an equally weighted manner, thereby, not involving attentive discrimination.

\textit{Our Contribution:} Given that stochmax improves intrinsic discrimination capability, we utilized this stochastic regularization effect to decrease the training/inference latency in SNN frameworks. We show how standard AGD can be integrated with stochmax classifier functionality to learn deep SNNs. Our analysis yields that deep SNNs of 3-4 layers trained with stochmax yield higher accuracy at lower latency than softmax baselines (for AGD training). 

\subsection{Addressing Accuracy with Network Hybridization}
It is evident that accuracy loss due to \textit{vanishing spike propagation} and \textit{input pixel-to-spike coding} are innate properties of SNN design that can be addressed to certain extent, but, cannot be completely eliminated. In order to achieve competitive accuracy as that of an ANN, we believe that taking a hybrid approach with a partly-artificial-and-partly-spiking neural architecture will be most beneficial. 


\textit{Our Contribution:} We demonstrate a hybrid neural architecture for AGD training methodologies. In case of AGD, since the training is performed end-to-end in a deep network, vanishing spike-propagation becomes a limiting factor to achieve high accuracy. To address this, we use ReLU based neurons in the initial layers and have spiking leaky-integrate-and-fire neurons in the latter layers and perform end-to-end AGD backpropagation. In this scheme, the idea is to extract relevant activity from the input in the initial layers with ReLU neurons. This allows the spiking neurons in latter layers to optimize the loss function and backpropagate gradients appropriately based on relevant information extracted from the input without any information loss. 


Finally, we show the combined benefits of incorporating BackRes connections with stochmax classifiers and network hybridization across different SNN training methodologies and show latency, accuracy and compute-efficiency gains. Through this work, our goal is to communicate good practices for deploying SNN frameworks that yield competitive performance and efficiency as compared to corresponding ANN counterparts. 

\section{SNN: Background and Fundamentals}
\subsection{Input and Neuron Representation}
\begin{figure}
\centering
\includegraphics[width=\linewidth]{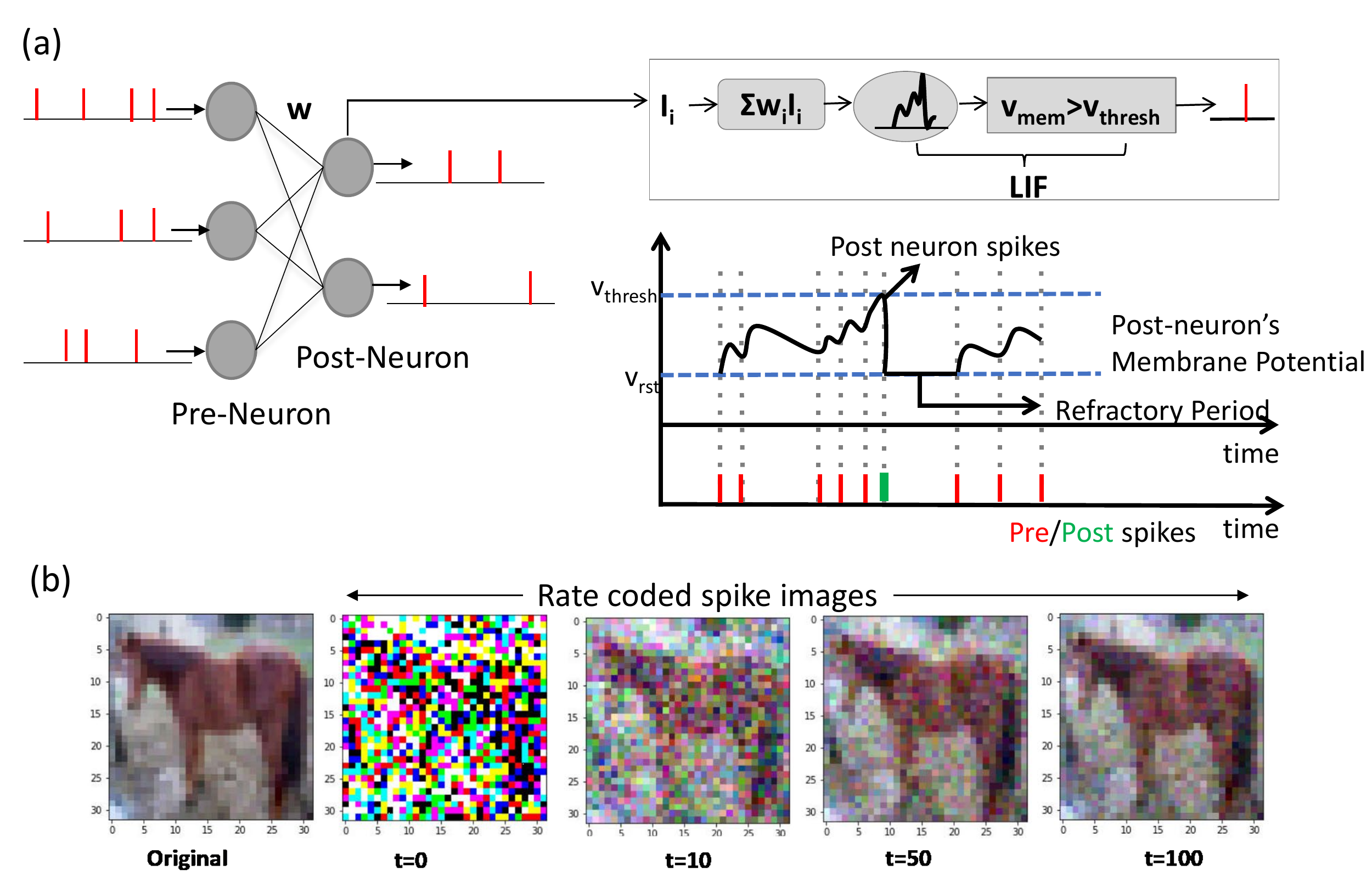}
\caption{(a) A feedforward fully-connected SNN architecture with Leaky-Integrate-and-Fire (LIF) spiking dynamics is shown. The notations correspond to Eqn. (1). (b) A sample CIFAR10 RGB pixel image (denoted as original) and corresponding rate-coded spike images at different time instants are shown. The spike image plotted at $t=n$ is a summation of all spike maps from $t=0$ to $t=n$.}
\end{figure}
Fig. 2(a) illustrates a basic spiking network architecture with Leaky-Integrate-and-Fire (LIF) neurons processing rate-coded inputs\footref{foot1}. 
It is evident from Fig. 2(b) that converting pixel values to binarized spike data \textit{\{1: spike, 0: no spike\}} in the temporal domain preserves the integrity of the image over several time steps. The dynamics of a LIF spiking neuron is given by
\begin{equation}
\tau \frac{d v_{mem}}{dt} = -v_{mem}+ \Sigma_i I_i w_i
\end{equation}

The membrane potential $v_{mem}$ integrates incoming spikes $I_i$ through weights $w_i$ and leaks (with time constant $\tau$) whenever it does not receive a spike. The neuron outputs a spike event when $v_{mem}$ crosses certain threshold $v_{thresh}$. Refractory period ensues after spike generation during which the post-neuron's membrane potential is not affected. In some cases, Integrate-and-Fire (IF) neurons are also used where leak value is 0 for simplicity in simulations/hardware implementations. 
Note, while Fig. 2 illustrates a fully-connected network, SNNs can be constructed with a convolutional hierarchy comprising multiple layers. For the sake of notation, we will refer to networks with real-valued computations/ReLU neurons as ANNs and networks with spike-based computations/LIF or IF neurons as SNNs.

\subsection{Training Methodology}
\subsubsection{Conversion from ANN-to-SNN}
To achieve higher accuracy with SNNs, a promising approach has been to convert ANNs trained with standard backpropagation into spiking versions. Fundamentally, the goal here is to match the input-output mapping function of the trained ANN to that of the SNN. Recent works \cite{diehl2015fast, sengupta2019going} have proposed weight normalization and threshold balancing methods in order to obtain minimal loss in accuracy during the conversion process. In this work, we use the threshold balancing method \cite{sengupta2019going} that yields almost zero-loss ANN-to-SNN conversion performance for deep VGG/ResNet-like architectures on complex Imagenet dataset.


In threshold balancing, after obtaining the trained ANN, the first step is to generate a Poisson spike train corresponding to the entire training dataset for a large simulation duration or time period (generally, \textit{2000 - 2500 time steps}). The Poisson spike train allows us to record the maximum summation of weighted spike input ($\Sigma_i w_i. X_i(t)$) received by the first layer of the ANN. $v_{thresh}$ value for the first layer is then set to the maximum summation value. After the threshold for the first layer is set, the network is again fed the input data to obtain a spike-train at the first layer, which serves as the input spike-stream for the second layer of the network. This process of generating spike train and setting $v_{thresh}$ value is repeated for all layers of the network. Note, the weights during this balancing process remain unchanged. For more details on this technique, please see \cite{sengupta2019going}.  

While conversion approach yields high accuracy, the computation cost is large due to high latency in processing. Reducing the time period from \textit{2000 to 100/10} time steps causes large decline in accuracy as $v_{thresh}$ balancing fails to match the output rate of SNN to that of ANN. 

\subsubsection{Approximate Gradient Descent (AGD)}
The thresholding functionality in the spiking neuron yields a discontinuous/non-differentiable functionality making it incompatible with gradient-descent based learning methods. Consequently, several training methodologies  have been proposed to incorporate the temporal statistics of SNNs and overcome the gradient descent challenges \cite{panda2016unsupervised, lee2016training, o2013real, lee2018deep, bellec2018long, neftci2019surrogate}. The main idea is to approximate the spiking neuron functionality with a continuously differentiable model or use surrogate gradients as a relaxed version of the real gradients to conduct gradient descent training. In our work, we use the surrogate gradient approach proposed in \cite{neftci2019surrogate}.  

In \cite{neftci2019surrogate}, the authors showed that temporal statistics incorporated in SNN computations can be implemented as a recurrent neural network computation graph (in, PyTorch, Tensorflow \cite{abadi2016tensorflow} frameworks) that can be unrolled to conduct Backpropagation Through Time (BPTT) \cite{werbos1990backpropagation}. The authors in \cite{neftci2019surrogate} also showed that using LIF computations in the forward propagation and surrogate gradient derivatives during backpropagation allows SNNs (of moderate depth) to be efficiently trained end-to-end. Using a recurrent computational graph enables the use of BPTT for appropriately assigning the gradients with chain rule in the temporal SNN computations. Here, for a given SNN, rate coded input spike trains are presented and the output spiking activity at the final layer (which is usually a softmax classifier) is monitored for a given time period. At the end of the time period, the loss from the final softmax layer is calculated and corresponding gradients are backpropagated through the unrolled SNN computation graph.

\begin{figure*}
\centering
\includegraphics[width=0.85\linewidth]{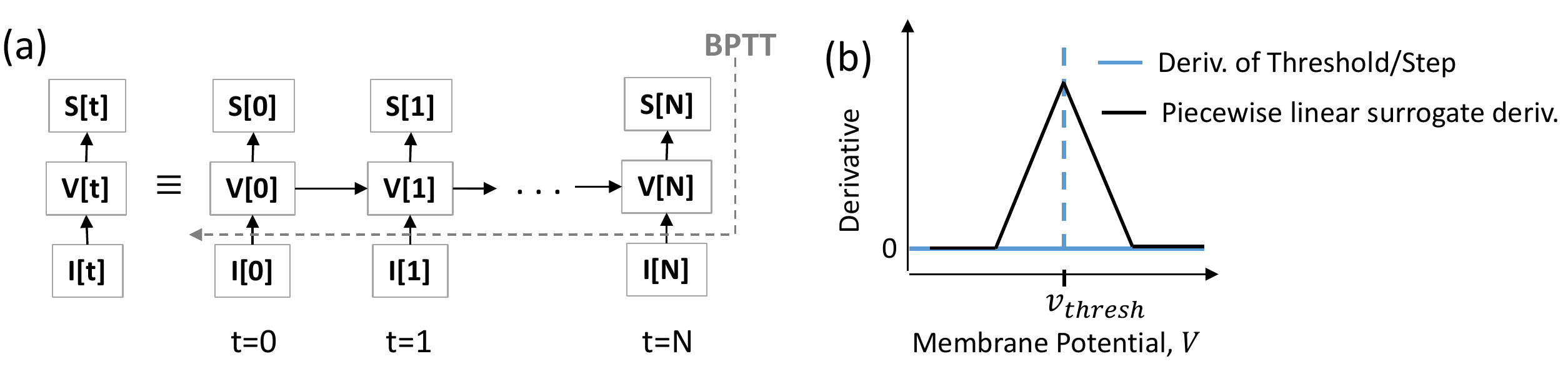}
\caption{(a) SNN computational graph unrolled over multiple time-steps for computing the $V$ or $V_{mem}$ and resultant output spike train $S$ as a function of input spikes $I[t]$ is shown. (b) Illustration of indeterminate derivative for a threshold or step function that is replaced with a surrogate piecewise linear derivative to allow the flow of gradients in AGD training.}
\end{figure*}

Fig. 3 (a) illustrates the SNN computational graph. From an implementation perspective, we can write the dynamics of an LIF neuron in discrete time as 
\begin{align}
V_{{mem}_i} [t+1] &= \alpha V_{{mem}_i}[t] + I_i[t] \\
I_i[t] &= \Sigma_j W_{ij} S_j[t]
\end{align}
Here, the output spike train $S_i$ of neuron $i$ at time step $t$ is a non-linear function of membrane potential $S_i[t] \equiv \Theta(V_{{mem}_i}[t] - v)$ where $\Theta$ is the Heaviside step function and $v$ is the firing threshold. $I_i$ is the net input current and $\alpha = exp(-\Delta_t/\tau_{mem}) $ is the decay constant (typically in the range $\{0.95,0.99\}$). During backpropagation, the derivative of $S(V_{mem} (t)) = \Theta(V_{mem} (t) - v)$ is zero everywhere except at $V_{mem} = v$ where it is not defined. This all-or-nothing behavior of spiking neurons stops gradient from flowing through chain rule making it difficult to perform gradient descent. We approximate the gradient using surrogate derivatives for $\Theta$ following \cite{neftci2019surrogate,bellec2018long} as 
\begin{equation}
\frac{dS[t]}{dV_{mem}[t]} = \gamma max\{0, 1- |\frac{V_{mem}[t] - v}{v}|\}
\end{equation}
where $\gamma$ is a damping factor (set to $0.3$) that yields stable performance during BPTT. As shown in Fig. 3 (b), using a surrogate gradient (Eqn. 4) now replaces a zero derivative with an approximate linear function. For more details and insights on surrogate gradient descent training, please see \cite{neftci2019surrogate, bellec2018long}. For convenience in notation, we will use AGD to refer to surrogate descent training in the remainder of the paper.


Using end-to-end training with spiking computations enables us to lower the computation time period to \textit{50 - 100} time steps. However, these methods are limited in terms of accuracy/performance and are also not suitable for training very deep networks.  

\subsubsection{Unsupervised STDP Learning}
STDP is a correlation based learning rule which modulates the weight between two neurons based on the correlation between pre- and post-neuronal spikes. In this work, we use a variant of the STDP model used in \cite{diehl2015unsupervised, srinivasan2018stdp,srinivasan2019restocnet} described as
\begin{equation}
\Delta w_{STDP} = \eta \times (e^{-(\frac{t_{post}-t{pre}}{\tau})} - STDP_{offset})
\end{equation}
where $\Delta w_{STDP} $ is the weight update, $\eta$ is the learning rate, $ t_{post}, t_{pre}$ are the time instants of post- and pre-neuronal spikes, $\tau$ is the STDP time constant. Essentially, the weight is increased if a pre-neuron triggers a post-neuron to fire within a time period specified by the $ STDP_{offset}$ implying strong correlation. If the spike timing difference is large between the pre- and post-neurons, the weight is decreased. In \cite{srinivasan2019restocnet}, the authors implemented a mini-batch version of STDP training for training convolutional SNNs in a layerwise manner. For training the weight kernels of the convolutional layers shared between the input and output maps, the pre-/post-spike timing differences are averaged across a given mini-batch and corresponding STDP updates are performed. In this work, we perform mini-batch training as in \cite{srinivasan2019restocnet, lee2018deep}. We also use the uniform threshold adaptation and dropout scheme following \cite{srinivasan2018stdp, lee2018deep, srinivasan2019restocnet} to ensure competitive learning with STDP. For more information on the learning rule, please see \cite{srinivasan2018stdp, lee2018deep}.

Generally, a network trained with layerwise STDP (for convolutional layers) is appended with a classifier (separately trained with backpropagation) to perform final prediction. 
The authors in \cite{srinivasan2019restocnet} showed that unsupervised STDP learning (even with binary/probabilistic weight updates) of a deep SNN, appended with a fully-connected layer of ReLU neurons, yields reasonable accuracy. However, similar to AGD, layerwise STDP training is not scalable and yields restrictive performance for deep multi-layered SNNs. 


\section{SNNs with BackRes Connections}
BackRes allows a model to perform complex computation over multiple logical depth by means of repeated unrolling. From Fig. 1, it appears that the number of output and input channels in a BackRes block need to be equal for consistency. However, given a BackRes block with 64 input channels and 128 output channels (say, VGG2x4 network), one can randomly drop 64 channels from the output during unrolled computations. Selecting top-64 channels with maximal activity, or averaging the response of 128 channels into 64 also yields similar accuracy as that of a baseline network (VGG8). For convenience, in our experiments, we use models with same input/output channels and convert them to BackRes blocks. Next, we discuss how to integrate BackRes connection for different SNN training methodologies.

\textit{Conversion :} In this methodology, SNN is constructed from a trained ANN. Hence, the ANN has to incorporate BackRes connections with repeated ReLU computations (similar to Fig. 1) which then need to be appropriately matched to spiking neuronal rates. Fig. 4 illustrates the conversion from ReLU to IF neurons. Here, since unrolling each time yields a different output rate, we need to ensure that we use multiple threshold IF neurons where $IF_{1}$ with threshold $v_{{thresh}_1}$ is activated for $n=1$ and $IF_{2}$ with threshold $v_{{thresh}_2}$ for $n=2$. Thus, the number of thresholds $v_{{thresh}_n}$ will be equal to the number of unrolling steps $n$. During threshold balancing for conversion (see Section III.B.1), we need to set the thresholds for each layer as well as each step of unrolling within a layer separately. Interestingly, we find that $v_{{thresh}}$ increases with $n$, i.e., $v_{{thresh}_1} < v_{{thresh}_2} … < v_{{thresh}_n}$. Increasing threshold implies lesser spiking activity with each unrolling which reduces the overall energy cost (results shown in Section VIII.A). 

\begin{figure}
\centering
\includegraphics[width=\linewidth]{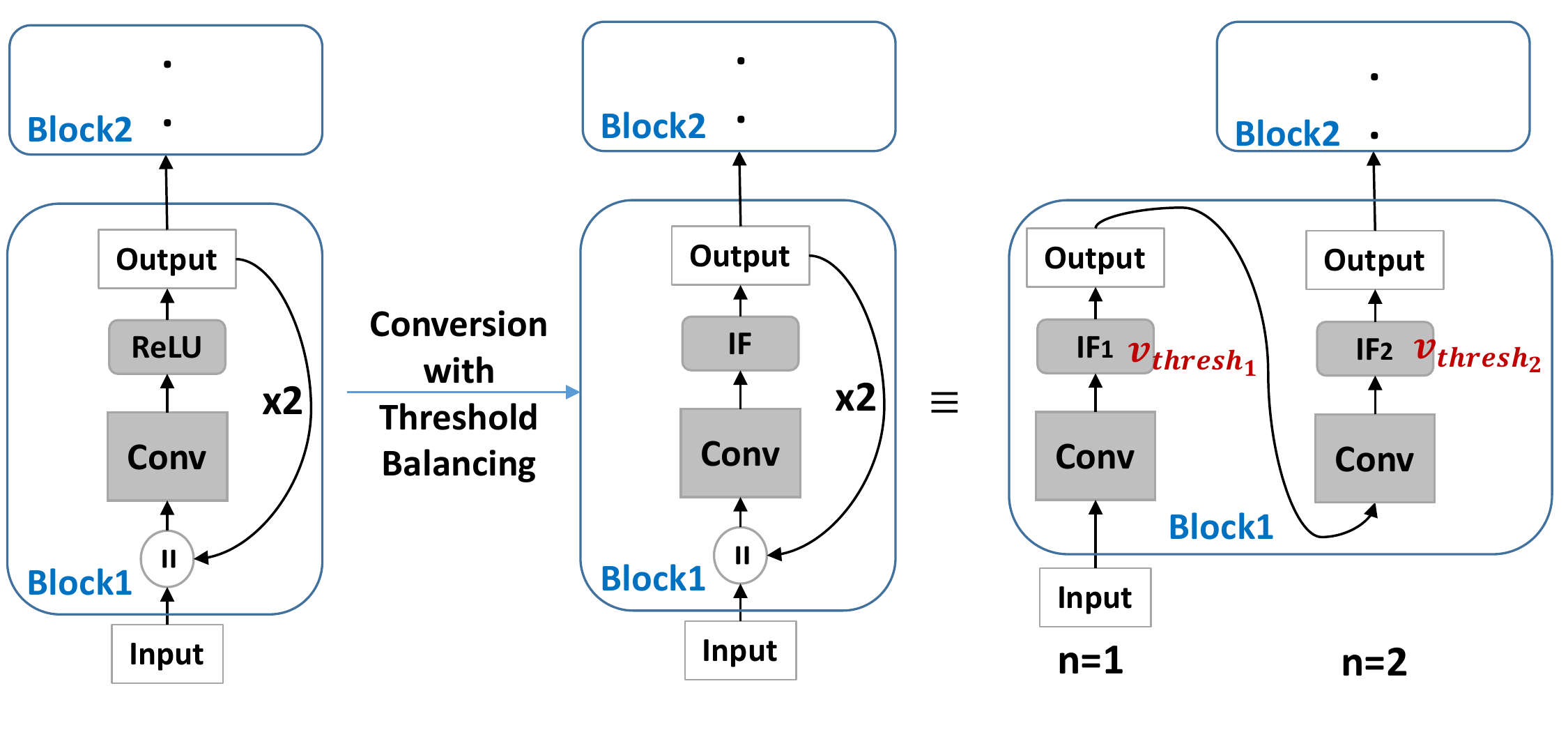}
\caption{Conversion of ANN with BackRes blocks into SNNs using threshold balancing. Here, during BackRes computations, we need to use multiple threshold $v_{{thresh}_{1,2}}$ IF neurons $IF_1, IF_2$ to match the input-output activity between ANN and SNN at each unrolling step $n=1, 2$.}
\end{figure}

\textit{AGD Training :} In AGD training, an SNN is trained end-to-end with the loss calculated at the output layer using surrogate gradient descent on LIF neurons. The thresholds of all neurons are set to a user-defined value at the beginning of training and remain constant throughout the learning process. The weight updates inherently account for the balanced spiking activity given the set thresholds. Adding BackRes blocks in this case will be similar to training a recurrent model with unrolled computation, that is treating the BackRes block as a feedforward network of $n$ layers. 
During backpropagation, the gradients are passed through the unrolled graph of the BackRes block, where, the same weights $w$ are updated $n$ times. 

\begin{figure}
\centering
\includegraphics[width=\linewidth]{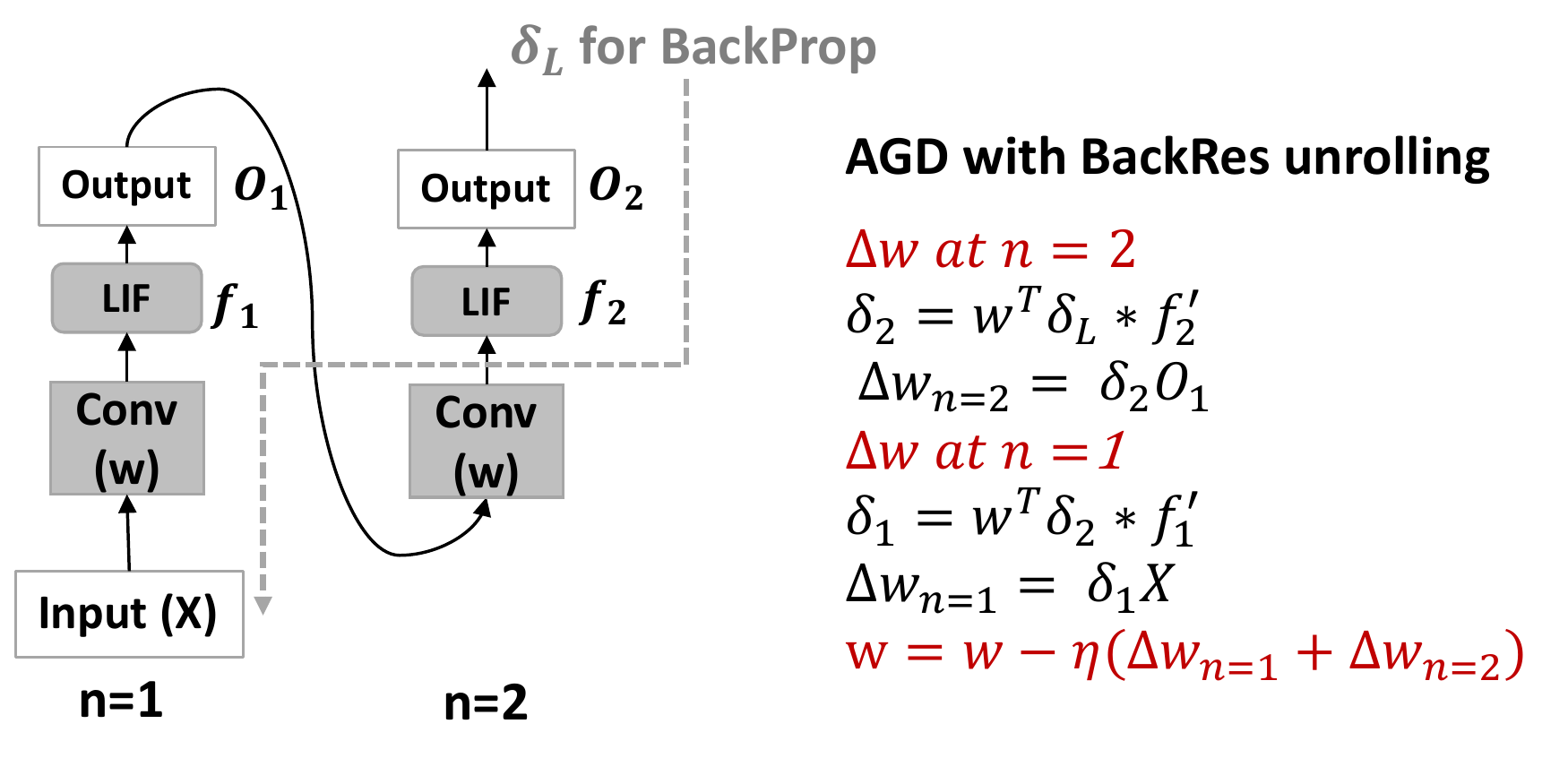}
\caption{AGD backpropagation chain rule update with BackRes connections. The BackRes block is essentially unrolled $n$ times and the loss is propagated through the unrolled graph to compute the weight updates at each unrolling step $n=1, 2$ as shown. Note, the $Conv$ layer weights $w$ of the BackRes block receive two updates with different input and output combinations giving rise to diverse gradients.}
\end{figure}

Fig. 5 illustrates the backpropagation chain rule update. It is worth mentioning that the LIF activity with every unrolling varies, that eventually affects the weight update value at each step. As in conversion, we find that networks with BackRes blocks and shared weights (say, VGG2x2) generally have lower spiking activity than equivalent depth baseline network with separate layers (say, VGG4), yielding energy improvements. This implies that the repeated computation with unrolling gives rise to diverse activity that can possibly model diverse features, thereby, allowing the network to learn even with lesser depth. Note, the BackRes network and the baseline network have same $v_{thresh}$ through all layers when trained with AGD. Further, AGD training has scalability limitations. For instance, a 7-layered VGG network fails to learn with end-to-end surrogate gradient descent. However, a network with BackRes blocks with real depth of 5 layers and logical depth of 7 layers can now be easily trained and in fact yields competitive accuracy (results shown in Section VIII.A). 

\textit{STDP Training :} SNNs learnt with STDP are trained layerwise in an iterative manner. Generally, in one iteration of training (comprising of $T$ time-steps or $1$ time period of input presentation), a layer’s weights are updated $k$ times ($k \le T$) depending upon the total spike activity in the pre-/post-layer maps and spiking correlation (as per Eqn. (5)). Since BackRes performs $n$ repeated computations of a single layer, in this case, we make $k \times n$ weight updates for the given layer in each iteration of STDP training. From Fig. 5, we can gather that the pre-/post- correlation at $n=1$ unrolling step will correspond to input $X$ and $Conv$ layer’s output that will determine its weight updates. For $n=2$, the $Conv$ layer’s output from previous step will serve as pre-spiking activity based on which the weights are updated again. Similar to AGD training, the overall activity at the output of $Conv$ changes with $n$ which diversifies and improves the capability of the network to learn better. We also find reduced energy cost and better scalability toward large logical depth networks that otherwise (with real depth) could not be trained in a layerwise manner (results shown in Section VIII.A). 

\section{SNNs with Stochmax}
Stochmax as noted earlier is a stochastic version of a softmax function that allows a network to discriminate better by focusing or giving importance to confusing classes. A softmax classifier is defined as
\begin{equation}
p(y|x;\theta) = \frac{exp(o_t(x;\theta))}{\Sigma_k exp(o_k(x;\theta))}
\end{equation}
where $t$ is the target label for input $x$, $k$ is the number of classes, and $o(x;\theta) = W^Th+b, \theta = \{W, b\}$ is the logits score generated from the last feature vector $h=NN(x;\omega)$ of a neural network $NN(.)$ parameterized by $\omega$. With Stochmax, the objective is to randomly drop out classes in the training phase with a motivation of learning an ensemble of classifiers in a single training iteration. From Eqn. (6), we can see that making $exp(o_k) = 0$ drops class $k$ completely even eliminating its gradients for backpropagation. Following this, Stochmax is defined as:
\begin{multline}
z_k|x \sim Ber(z_k;\rho_k(x;\theta), \\
p(y|x,z;\theta,\psi) = \frac{(z_t + \epsilon)exp(o_t(x;\theta))}{\Sigma_k (z_k + \epsilon) exp(o_k(x;\theta))}
\end{multline}
Here, we drop out classes with a probability ($1-\rho_k$) based on Bernoulli ($Ber$) trials. Further, to encode meaningful correlations in the probabilities $\rho_k$, we learn the probabilities as an output of the neural network which takes last feature vector $h$ as input and outputs a sigmoidal value $\rho(x;\psi) = \sigma({{W^T}_{\psi}} + b_\psi), \psi = \{W_\psi, b_\psi\}$. By learning $\psi$, we expect that highly correlated classes can be dropped or retained together. In essence, by dropping classes, we let the network learn on different \textit{sub-problems} at each iteration. 
In SNN implementations, we replace the softmax classifier (Eqn. (6)) with a Stochmax function (Eqn. (7)) at the output. Generally, the classifier layer is a non-spiking layer which receives accumulated input from the previous spiking layer $h$ integrated over the $T$ time-steps per training iteration. The loss is then calculated from stochmax output which is used to calculate the gradients and perform weight updates. 

It is evident that AGD training where the loss function at the classifier is used to update the weights at all layers of a deep SNN will be affected by this softmax-to-stochmax replacement. We find that this attentive discrimination that implicitly models many classifiers (providing different decision boundaries) per training iteration allows an SNN to be trained even with lower latency (or lesser time steps per training iteration or input presentation) while yielding high accuracy. Lower latency implies that pixel-to-spike input coding with Poisson rate will incur more loss. However, the deficit of the input coding gets rectified with improved classification.

In Conversion, an ANN is trained separately and is completely dissociated from the spiking statistics. STDP, on similar lines, has spike-based training of intermediate feature extractor layers. The final classifier layers (which are separately trained) are appended to the STDP-trained layers and again do not influence the weight or activity learnt in the previous layers. Thus, while Stochmax classifier inclusion slightly improves the accuracy of both conversion/STDP methods, they remain unaffected from a latency perspective. 

\section{SNNs with Hybrid ReLU-and-LIF neurons}
The objective with a partly-artificial-and-partly-spiking neural architecture is to achieve improved accuracy. For artificial-to-spiking conversion methodology, since training is performed using ReLU neuronal units and inference with spiking integrate-and-fire neurons, network hybridization is not necessary and will not add to the overall accuracy. Most works on STDP learning use hybrid network architecture where STDP is used to perform feature extraction with greedy layer-wise training of the convolutional layers of a deep network. Then, a one-layer fully connected ANN (with ReLU neurons) is appended to the STDP trained layers to perform final classification. However, STDP is limited in its capability to extract specific features from the input that are key for classification. We find that strengthening the ANN hierarchy of an STDP-trained SNN (either with Stochmax or deepening the ANN with multiple layers) yields significant improvement in accuracy.

In AGD, since learning is performed end-to-end vanishing spike-propagation restricts the training of a deep many-layered network. For instance, a VGG7 network fails to train with AGD. In fact, even with residual or skip connections that leads to a ResNet7-like architecture, the model is difficult to train. BackRes connections are potential solutions for training logically deep networks. However, to achieve better accuracy for deeper many-layered networks, there is a need to hybridize the layers of the network with ReLU and LIF neurons. 

\begin{figure*}
\centering
\includegraphics[width=0.75\linewidth]{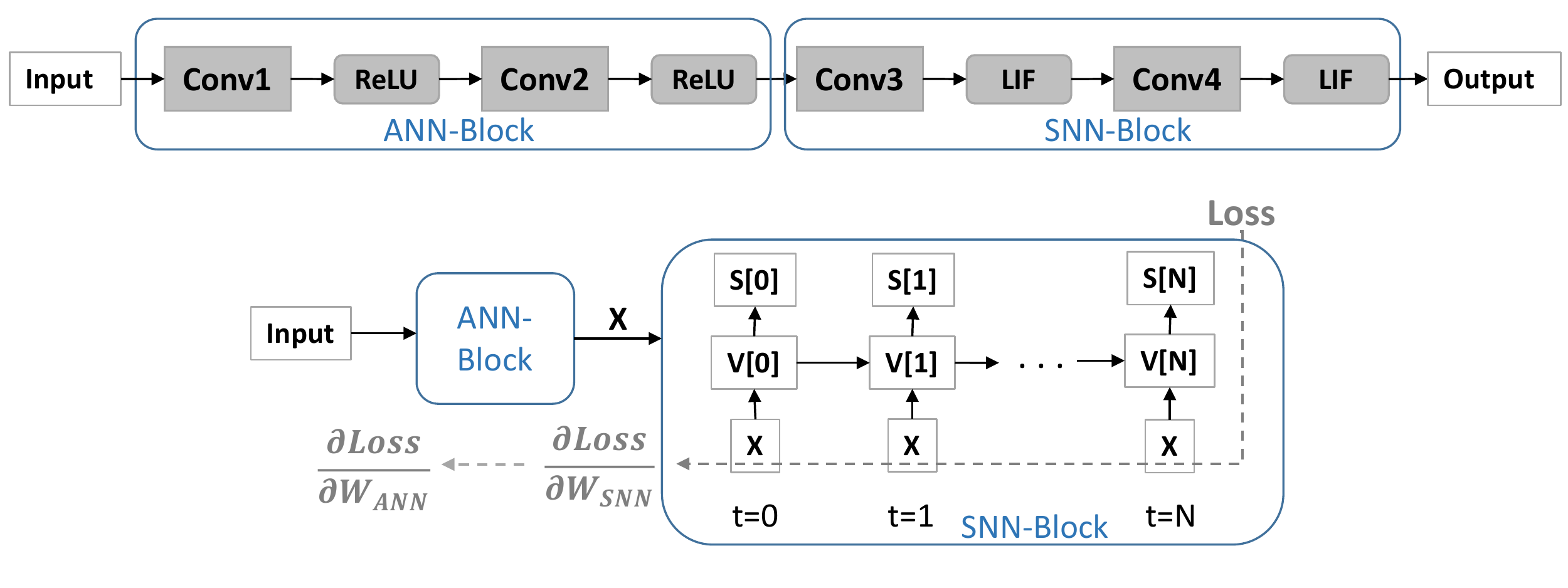}
\caption{A hybrid network architecture with ReLU activation initially and LIF activation in latter layers for AGD training is shown. During AGD backpropagation, the $SNN-block$ is unrolled and the weight updates are calculated through the unrolled graph over different time-steps using chain-rule. The loss gradient from the $SNN-block$, $\frac{\partial Loss}{\partial W_{SNN}}$, is then backpropagated through the ANN computational graph to calculate the ANN loss gradients $\frac{\partial Loss}{\partial W_{ANN}}$. Note, the $SNN-block$ recieves the real-valued input $X$ computed from the $ANN-block$ at each time-step.}
\end{figure*}

Fig. 6 illustrates the hybrid network configuration. We have ReLU neurons in initial layers and temporal LIF neurons in latter layers. During forward propagation, the input processed through the $ANN-block$ is then propagated through the $SNN-block$ unrolled over different time-steps as a recurrent computational graph to calculate the loss at the final output layer (that can be softmax/stochmax function). In the backward phase, the gradient of loss is propagated through the recurrent graph updating the weights of the SNN block with surrogate linear approximation of the LIF functionality corresponding to activity at each time step. The loss gradient calculated through BPTT are then passed through the ANN-block (which calculates the weight updates in ANN with standard chain rule). It is worth mentioning that setting up a hybrid network in a framework like PyTorch automatically performs recurrent graph unrolling for SNN-block and standard feedforward graph for ANN-block and enables appropriate gradient calculation and weight updates. We would also like to note that we feed in the output of the ANN-block as it is (without any rate-coding) to the SNN-block. That is, the unrolled SNN graph at each time-step receives the same real-valued input $X$. We find that processing $X$ instead of rate-coded $X[t]$ yields higher accuracy at nearly-same energy or computation cost. 

\section{Experiments}
We conduct a series of experiments for each optimization scheme, primarily, using CIFAR10 and Imagenet data on VGG and ResNet architectures of different depths detailing the advantages and limitation of each approach. We implemented all SNN models in PyTorch and used the same hyperparameters (such as, leak time constant, $v_{thresh}$ value, input spike rate etc.) as used in \cite{sengupta2019going}, \cite{neftci2019surrogate}, \cite{srinivasan2019restocnet} for conversion, surrogate gradient descent and STDP training, respectively. In all experiments, we measure the accuracy, latency, energy or total compute cost, and total parameters for a given SNN implementation and compare it to the baseline ANN counterpart. Latency is measured as total number of time-steps required to perform an inference for one input. In case of ANN, latency during inference is 1, while, SNN latency is the total number of time-steps $T$ over which an input is presented to the network. Note, in all our experiments, all ANNs and SNNs are trained for different number of epochs/iterations until maximum accuracy is achieved in each case.

\begin{table}
\caption{\textbf{Energy table for 45nm CMOS process.}}
\label{table1}
\centering
\begin{tabular}{|c|c|p{\textwidth}}
\hline
Operation&Energy (pJ) \\
\hline
32-b MULT Int & 3.1\\
32-b ADD Int & 0.1\\
32-b MAC Int & 3.2 ($E_{MAC}$)\\
32-b AC Int & 0.1 ($E_{AC}$)\\
\hline
\end{tabular}
\end{table}

The total compute cost is measured as total number of floating point operations (FLOPS) which is roughly equivalent to the number of multiply-and-accumulate (MAC) or dot product operations performed per inference per input \cite{han2015learning, han2015deep}. In case of SNN, since the computation is performed over binary events, only accumulate (AC) operations are required to perform the dot product (without any multiplier). Thus, SNN /ANN FLOPS count will consider AC/MAC operations, respectively. For a particular convolutional layer of an ANN/SNN, with $N$ input channels, $M$ output channels, input map size $I\times I$ , weight kernel size $k\times k$ and output size $O\times O$, total FLOPS count for ANN/SNN is 
\begin{align}
FLOPS_{ANN} &= O^2 * N * k^2 * M\\
FLOPS_{SNN} &= O^2 * N * k^2 * M * S_A
\end{align}
Note, $FLOPS_{SNN}$ in Eqn. (9) is calculated per time-step and considers the net spiking activity ($S_A$) that is the total number of firing neurons per layer. In general, $S_A <<1$ in an SNN on account of sparse event-driven activity, whereas, in ANNs $S_A=1$. For energy calculation, we specify each MAC or AC operation at the register transfer logic (RTL) level for 45nm CMOS technology \cite{han2015learning}. Considering 32-bit weight values, the energy consumption for a 32-bit integer MAC/AC operation ($E_{MAC}, E_{AC}$) is shown in Table \ref{table1}. Total inference energy $E$ for ANN/SNN considering FLOPS count across all $N$ layers of a network is defined as
\begin{align}
E_{ANN} &= (\sum_{i=1}^{N}FLOPS_{ANN}) * E_{MAC}\\
E_{SNN} &= (\sum_{i=1}^{N}FLOPS_{SNN}) * E_{AC} *T
\end{align}
For SNN, the energy calculation considers the latency incurred as the rate-coded input spike train has to be presented over $T$ time-steps to yield the final prediction result. Note, this calculation is a rather rough estimate which does not take into account memory access energy and other hardware architectural aspects such as input-sharing or weight-sharing. Given, that memory access energy remains same irrespective of SNN or ANN network topology, the overall \textit{Energy-Efficiency (EE)} $EE = E_{ANN}/E_{SNN}$ will remain unaffected with or without memory access consideration. Finally, to show the advantage of utilizing BackRes connections, we also compute the total number of unique parameters (i.e. total number of weights) in a network and calculate the compression ratio that BackRes blocks yield over conventional feedforward blocks of similar logical depth.

\section{Results}
\subsection{Impact of BackRes Connections}
\begin{table}
\caption{\textbf{Accuracy \& Total \# parameters for ANN and corresponding converted SNN topologies (refer Table \ref{table3})  for different latency $T$ on CIFAR10 data}}
\label{table2}
\centering
\begin{tabular}{|c|c|c|c|c|p{\textwidth}}
\hline
Model&\makecell{ANN\\($T=1$)}&\makecell{SNN\\($T=250$)}&\makecell{SNN\\($T=2500$)} & \#Parameters\\
&\multicolumn{3}{|c|}{(---------Accuracy (\%)---------)}&\\
\hline
VGG7 & 88.74 & 85.88 & 88.56 & 1.2M (1x)\\
VGG2x4 & 86.14 & 81.99 & 86.23 & 1.09M (1.1x)\\
VGG3x2 & 87.34 & 83.31 & 87.15 & 1.13M (1.06x)\\
\hline
\end{tabular}
\end{table}

\begin{table}
\caption{\textbf{CIFAR10 Network Topologies for Conversion training. ConvN(I,O,k$\times$k/s) denotes $N^{th}$ convolutional layer with $I$ input channels, $O$ output channels, kernel of size $k \times k$ with stride $s$. Pool(p$\times$p/$s_p$) denotes average pooling layer with pooling window size $p\times p$ and pooling stride $s_p$. FC(X,Y) denote a fully-connected layer with $X$ input nodes and $Y$ output nodes. Layers with BackRes connections and repeated computations have been highligted in \textcolor{red}{red}.}}
\label{table3}
\centering
\begin{tabular}{|c|c|c|p{\textwidth}}
\hline
Model & Configuration & BackRes\\
\hline
VGG7 & Input--Conv1(3,64,3x3/1)--Conv2(64,64,3x3/1)-- & \multirow{4}{4em}{Not Applicable}\\
&--Conv3(64,64,3x3/1)-- Conv4(64,64,3x3/1)-- &\\
&--Conv5(64,64,3x3/1)--Pool(2x2/2)--Pool(2x2/2)--&\\
&--Pool(2x2/2)--FC1(2048,512)--FC2(512,10)&\\

\hline
VGG2x4 & Input--Conv1(3,64,3x3/1)--\textcolor{red}{Conv2(64,64,3x3/1)}-- & \multirow{4}{4em}{\textcolor{red}{[Conv2]} repeated 4 times}\\
&--\textcolor{red}{Conv2(64,64,3x3/1)}--\textcolor{red}{Conv2(64,64,3x3/1)}-- & \\
&--\textcolor{red}{Conv2(64,64,3x3/1)}--Pool(2x2/2)--Pool(2x2/2)--&\\
&--Pool(2x2/2)--FC1(2048,512)--FC2(512,10)&\\
\hline
VGG3x2 & Input--Conv1(3,64,3x3/1)--\textcolor{red}{Conv2(64,64,3x3/1)}--&  \multirow{4}{4em}{\textcolor{red}{[Conv2--Conv3]} repeated 2 times}\\
&--\textcolor{red}{Conv3(64,64,3x3/1)}--\textcolor{red}{Conv2(64,64,3x3/1)}-- &\\
&--\textcolor{red}{Conv3(64,64,3x3/1)}--Pool(2x2/2)--Pool(2x2/2)--&\\
&--Pool(2x2/2)--FC1(2048,512)--FC2(512,10)&\\
\hline
\end{tabular}
\end{table}

\begin{table}
\caption{\textbf{Accuracy, Total \# parameters and Energy Efficiency $EE$ for converted SNN topologies (refer Table \ref{table5}) of latency $T=2500$ and corresponding ANN on Imagenet data}}
\label{table4}
\centering
\begin{tabular}{|c|c|c|c|c|p{\textwidth}}
\hline
Model&\makecell{ANN\\($T=1$)}&\makecell{SNN\\($T=2500$)} & \makecell{\#\\Parameters}& \makecell{$EE =$ \\ $\frac{E_{ANN}(1\times)}{E_{SNN}}$}\\
&\multicolumn{2}{c|}{(Accuracy (Top-1/\textcolor{red}{Top-5}\%))}&&\\
\hline
VGG16 & \makecell{70.52/\\ \textcolor{red}{89.39}} & \makecell{69.96/\\ \textcolor{red}{89.01}} & \makecell{123.8M\\ (1x)} & 1.975x\\
\hline
\makecell{VGG11x2} & \makecell{69.72/\\ \textcolor{red}{88.56}} & \makecell{68.57/\\ \textcolor{red}{87.66}} & \makecell{116.1M\\ (1.07x)} & 3.66x\\
\hline
\end{tabular}
\end{table}

\begin{table*}
\caption{\textbf{Imagenet Network Topologies for Conversion training. Notations are same as that of Table \ref{table3}. Layers with BackRes connections and repeated computations have been highligted in \textcolor{red}{red}.}}
\label{table5}
\centering
\begin{tabular}{|c|c|c|p{\textwidth}}
\hline
Model & Configuration & BackRes\\
\hline
VGG16 & Input--Conv1(3,64,3x3/1)--Conv2(64,64,3x3/1)-- & \multirow{6}{4em}{Not Applicable}\\
&--Pool(2x2/2)--Conv3(64,128,3x3/1)--Pool(2x2/2)--Conv4(128,256,3x3/1)--&\\
&--Conv5(256,256,3x3/1)--Conv6(256,256,3x3/1)--Pool(2x2/2)--Conv7(256,512,3x3/1)--&\\
&--Conv8(512,512,3x3/1)--Conv9(512,512,3x3/1)--Conv10(512,512,3x3/1)--Conv11(512,512,3x3/1)--&\\
&--Conv12(512,512,3x3/1)--Conv13(512,512,3x3/1)--Pool(2x2/2)--Pool(2x2/2)--&\\
&--FC1(25088,4096)--FC2(4096,1000)&\\

\hline
VGG11x2 & Input--Conv1(3,64,3x3/1)--Conv2(64,64,3x3/1)-- & \multirow{6}{4em}{\textcolor{red}{[Conv5]}\& \textcolor{red}{[Conv7-Conv8-Conv9]} repeated 2 times}\\
&--Pool(2x2/2)--Conv3(64,128,3x3/1)--Pool(2x2/2)--Conv4(128,256,3x3/1)--\textcolor{red}{Conv5(256,256,3x3/1)}--&\\
&--\textcolor{red}{Conv5(256,256,3x3/1)}--Pool(2x2/2)--Conv6(256,512,3x3/1)--\textcolor{red}{Conv7(512,512,3x3/1)}--&\\
&--\textcolor{red}{Conv8(512,512,3x3/1)--Conv9(512,512,3x3/1)}--\textcolor{red}{Conv7(512,512,3x3/1)--Conv8(512,512,3x3/1)}--&\\
&--\textcolor{red}{Conv9(512,512,3x3/1)}--Pool(2x2/2)--Pool(2x2/2)--&\\
&--FC1(25088,4096)--FC2(4096,1000)&\\
\hline
\end{tabular}
\end{table*}

\begin{table}
\setlength{\tabcolsep}{4pt}
\caption{\textbf{Accuracy, Total \# parameters and Energy Efficiency $EE$ for STDP-trained SNN topologies (refer Table \ref{table7})  of latency $T=100$ and corresponding ANN on CIFAR10 data. $EE_{Conv}$ considers the energy calculated only for the convolutional/pooling layers excluding the FC layers, $EE_{Full}$ considers the total energy of the network including the FC layers.}}
\label{table6}
\centering
\begin{tabular}{|c|c|c|c|c|p{\textwidth}}
\hline
Model&\makecell{ANN\\($T=1$)}&\makecell{SNN\\($T=100$)} & \makecell{\#\\Parameters}& \makecell{$EE_{Conv} / EE_{Full} =$ \\ $\frac{E_{ANN}(1\times)}{E_{SNN}}$}\\
&\multicolumn{2}{|c|}{(--Accuracy\%--)}&&\\
\hline
ResNet2 & 78.26 & 61.02 & 18.9M & 1.64x/1.16x\\
ResNet3 & 80.11 & 51.1 & 28.37M & 1.81x/1.28x \\
ResNet2x2 & 79.39 & 63.21 & 28.35M & 10.56x/1.78x\\
\hline
\end{tabular}
\end{table}

\begin{table*}
\caption{\textbf{CIFAR10 Network Topologies for STDP training methodology. Notations are same as that of Table \ref{table3}. Layers with BackRes connections and repeated computations have been highligted in \textcolor{red}{red}. Forward Residual or Skip connections between layers of a network are denoted in  \textcolor{blue}{blue}.}}
\label{table7}
\centering
\begin{tabular}{|c|c|c|c|p{\textwidth}}
\hline
Model & Configuration & BackRes & Skip\\
\hline
ResNet2 & Input--Conv1(3,36,3x3/1)--Conv2(36,36,3x3/1)-- & \multirow{2}{4em}{Not Applicable} & \multirow{2}{10em}{\textcolor{blue}{Input-to-Conv2, Conv1-to-FC1}}\\
&--Pool(2x2/2)--FC1(18432,1024)--FC2(1024,10)&&\\
\hline
ResNet3 & Input--Conv1(3,36,3x3/1)-- & \multirow{4}{4em}{Not Applicable} & \multirow{4}{10em}{\textcolor{blue}{Input-to-Conv2, Conv1-to-FC1, Conv2-to-FC1}}\\
&--Conv2(36,36,3x3/1)--&&\\
&--Conv3(36,36,3x3/1)--Pool(2x2/2)--&&\\
&--FC1(27648,1024)--FC2(1024,10)&&\\
\hline
ResNet2x2 & Input--Conv1(3,36,3x3/1)-- & \multirow{4}{4em}{\textcolor{red}{[Conv2]} repeated 2 times} & \multirow{4}{10em}{\textcolor{blue}{Input-to-Conv2, Conv1-to-FC1}}\\
&--\textcolor{red}{Conv2(36,36,3x3/1)}--&&\\
&--\textcolor{red}{Conv2(36,36,3x3/1)}--Pool(2x2/2)--&&\\
&--FC1(18432,1024)--FC2(1024,10)&&\\
\hline
\end{tabular}
\end{table*}
First, we show the impact of incorporating BackRes Connections for conversion based SNNs. Table \ref{table2} compares the accuracy and total \# parameters across different network topologies (described in Table \ref{table3}) for ANN/SNN implementations on CIFAR10 data. For the sake of understanding, we provide the unrolled computation graph of networks with BackRes blocks and repeated computations in Table \ref{table3}. For instance, VGG2x4 refers to a network which has two unique convolutional layers ($Conv1, Conv2$) where $Conv2$ receives a BackRes Connection from its output and is computed $4$ times before processing the next layer as depicted in Table \ref{table3}. Similarly, VGG3x2 refers to a network with 3 unique convolutional layers ($Conv1, Conv2, Conv3$) with $Conv2, Conv3$ computation repeated 2 times in the order depicted in Table \ref{table3}. Note, VGG2x4/VGG3x2 achieve the same logical depth of a 7-unique layered (including fully connected layers) VGG7 network.

In Table \ref{table2}, we observe that accuracy of ANNs with BackRes connections suffer minimal loss (upto $\sim1-2\%$ loss) to that of the baseline ANN-VGG7 model. The corresponding converted SNNs with BackRes connections also yield near-accuracy. It is evident that SNNs with higher computation time or latency $T$ yield better accuracy.  While the improvement in total \# parameters is minimal here, we observe a significant improvement in energy efficiency ($EE = \frac{E_{ANN}(1\times)}{E_{SNN}}$ calculated using Eqn. (10), (11)) with BackRes additions as shown in Fig. 7. Note, the $EE$ of SNNs shown in Fig. 7 is plotted by taking the corresponding ANN topology as baseline ($EE$ of VGG2x4 SNN is measured with respect to VGG2x4 ANN). The large efficiency gains observed can be attributed to the sparsity obtained with event-driven spike-based processing as well as the repeated computation achieved with BackRes connections. In fact, we find that net spiking activity for a given layer decreases over repeated computations (implying a \textit{`sparsifying effect'}) with each unrolling step (due to increasing threshold per unrolling, see Section IV). Consequently, VGG2x4 with $n=4$ repeated computation yields larger $EE$ ($\sim1.3 \times$) than VGG3x2 ($n=2$).

\begin{figure}
\centering
\includegraphics[width=0.85\linewidth]{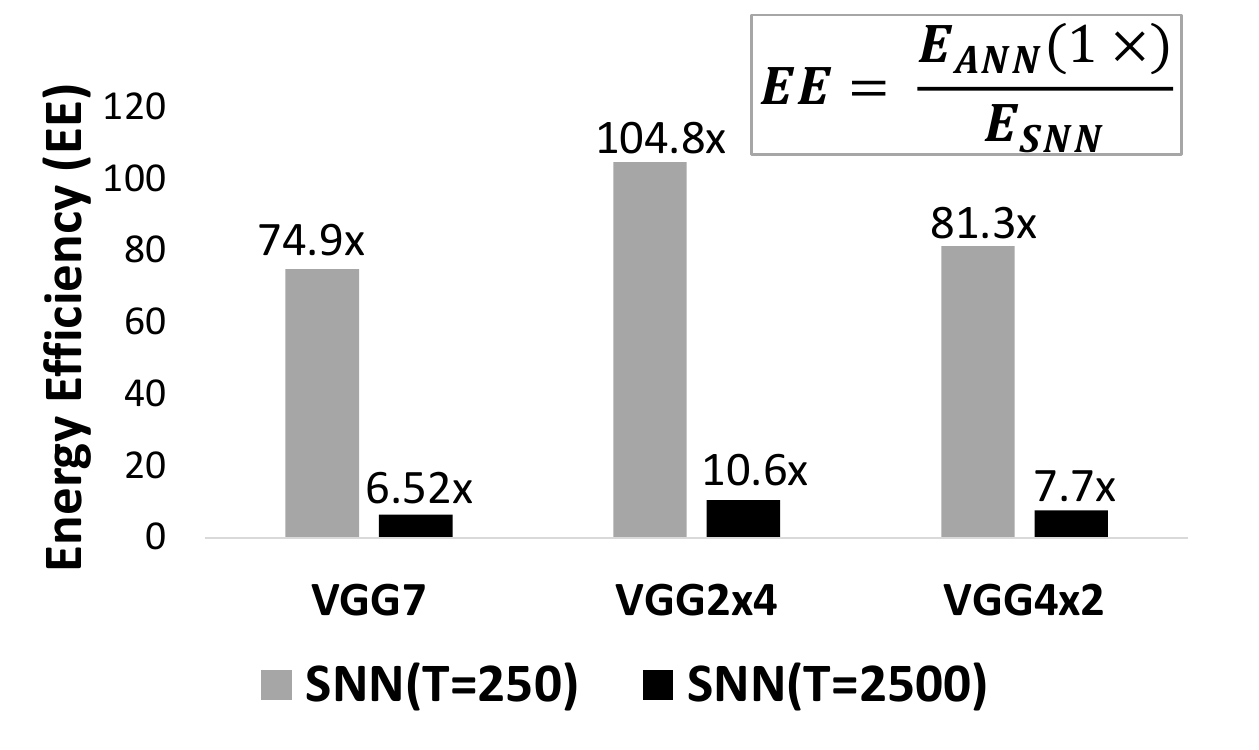}
\caption{Energy-Efficiency $EE$ results for different SNN topologies (from Table \ref{table3}) with/without BackRes connections trained with Conversion technique on CIFAR10 data. The efficiency values have been denoted on top of each graph for clarity. Note, $EE >1$ implies $E_{ANN} > E_{SNN}$ denoting lower energy consumption with SNN implementations.}
\end{figure}

Table \ref{table4} illustrates the Top-1/Top-5 accuracy, parameter compression ratio and $EE$ benefits observed with BackRes connections on Imagenet dataset (for topologies shown in Table \ref{table5}). Note, VGG11x2 (comprising of 11 unique convolutional or fully-connected layers) with BackRes connections and repeated computations achieves the same logical depth of 16 layers as that of VGG16. The accuracy loss in VGG11x2 (SNN) is minimal $\sim 1\%$ while yielding $\sim 2 \times$ greater $EE$ compared to VGG16 (SNN). We also find that for complex datasets like Imagenet, lower latency of $T=250$ yields very low accuracy with or without BackRes computations.

Next, we evaluate the benefits of adding BackRes connections for SNNs trained with STDP. As discussed earlier, in STDP training, the convolutional layers of a network are trained layerwise with LIF neurons. Then, an ANN classifier is appended to the STDP trained layers, wherein, the ANN classifier is trained separately on the overall spiking activity collected at the SNN layers. Table \ref{table6} shows the accuracy, \# parameters and $EE$ benefits of SNN topologies (listed in Table \ref{table7}) with respect to corresponding ANN baselines. All ANN baselines are trained end-to-end with backpropagation and requires the entire CIFAR10 training dataset (50,000 labelled instances). On the other hand, all SNNs requires only 5000 instances for training the Convolutional layers. Then, the fully-connected classifier (comprising of $FC1, FC2$ layers in Table \ref{table7}) appended separately to the STDP-learnt layers are trained on the entire CIFAR10 dataset. 
\begin{table}
\setlength{\tabcolsep}{4pt}
\caption{\textbf{Accuracy, Total \# parameters and Energy Efficiency $EE$ for AGD trained SNN topologies (refer Table \ref{table9}) of latency $T=25, 50$ and corresponding ANN on CIFAR10 data.}}
\label{table8}
\centering
\begin{tabular}{|c|c|c|c|c|c|p{\textwidth}}
\hline
Model&\makecell{ANN\\($T=1$)}&\makecell{SNN\\($T=25$)} &\makecell{SNN\\($T=50$)} &\makecell{\#\\Parameters}& \makecell{$EE =$ \\ $\frac{E_{ANN}(1\times)}{E_{SNN} (T=25)}$}\\
&\multicolumn{3}{|c|}{(------Accuracy\%------)}&&\\
\hline
VGG5 & 75.86 & 71.92 & 72.77 & 2.21M & 14.75x\\
VGG3x2 & 74.99 & 71.07 & 71.97 & 2.18M & 16.2x \\
\hline
VGG7 & 72.26 & - & - & 2.3M & -\\
VGG3x4 & 69.52 & 74.23 & 75.01 & 2.19M & 26.44x\\
\hline
\end{tabular}
\end{table}

\begin{table*}
\caption{\textbf{CIFAR10 Network Topologies for AGD training methodology. Notations are same as that of Table \ref{table3}. Layers with BackRes connections and repeated computations have been highligted in \textcolor{red}{red}.}}
\label{table9}
\centering
\begin{tabular}{|c|c|c|c|p{\textwidth}}
\hline
Model & Configuration & BackRes\\
\hline
VGG5 & Input--Conv1(3,64,3x3/1)--Conv2(64,64,3x3/1)-- & \multirow{2}{4em}{Not Applicable} \\
&--Pool(2x2/2)--Conv3(3,64,3x3/1)--Conv4(64,64,3x3/1)--&\\
&--Pool(2x2/2)--FC1(4096,512)--FC2(512,10)&\\
\hline
VGG3x2 & Input--Conv1(3,64,3x3/1)--Conv2(64,64,3x3/1)-- & \multirow{3}{4em}{\textcolor{red}{[Conv3]} repeated 2 times} \\
&--Pool(2x2/2)--\textcolor{red}{Conv3(3,64,3x3/1)--Conv3(64,64,3x3/1)}--&\\
&--Pool(2x2/2)--FC1(4096,512)--FC2(512,10)&\\
\hline
VGG7 & Input--Conv1(3,64,3x3/1)--Conv2(64,64,3x3/1)-- & \multirow{2}{4em}{Not Applicable} \\
&--Pool(2x2/2)--Conv3(3,64,3x3/1)--Conv4(64,64,3x3/1)--&\\
&--Conv5(3,64,3x3/1)--Conv6(64,64,3x3/1)--&\\
&--Pool(2x2/2)--FC1(4096,512)--FC2(512,10)&\\
\hline
VGG3x4 & Input--Conv1(3,64,3x3/1)--Conv2(64,64,3x3/1)-- & \multirow{4}{4em}{\textcolor{red}{[Conv3]} repeated 4 times} \\
&--Pool(2x2/2)--\textcolor{red}{Conv3(3,64,3x3/1)--Conv3(64,64,3x3/1)}--&\\
&--\textcolor{red}{Conv3(3,64,3x3/1)--Conv3(64,64,3x3/1)}--&\\
&--Pool(2x2/2)--FC1(4096,512)--FC2(512,10)&\\
\hline
\end{tabular}
\end{table*}
From Table \ref{table6}, we observe that SNN accuracy is considerably lower than corresponding ANN accuracy. This can be attributed to the limitation of STDP training to extract relevant features in an unsupervised manner. In fact, deepening the network from ResNet2 to ResNet3 causes a decline in accuracy corroborating the results of previous works \cite{srinivasan2019restocnet}. However, adding BackRes connection in ResNet2x2 which achieves same logical depth as ResNet3 improves the accuracy of the network while yielding significant gains ($\sim 10 \times$) in terms of $EE$. For $EE$, we show the gains considering the full network $EE_{Full}$ (including spiking convolutional and ReLU FC layers), as well as, the gain considering only the spiking convolutional layers $EE_{Conv}$. The spiking layers on account of event-driven sparse computing exhibit higher efficiency than the full network (i.e. $EE_{Conv} > EE_{Full}$). Interestingly, ResNet2x2 yields $\sim10 \times$ higher efficiency at the spiking layers which further supports the fact that BackRes connections have a \textit{`sparsifying'} effect on the intrinsic spiking dynamics of the network. This result establishes the advantage of BackRes connection in enabling scalability of STDP-based SNN training methodologies towards larger logical depth while yielding both accuracy and efficiency improvements.

For AGD training, BackRes additions yield both accuracy and scalability related benefits. Table \ref{table8} shows the accuracy, \# parameters and $EE$ benefits of SNN topologies (listed in Table \ref{table9}) for different latency $T=25,50$ with respect to corresponding ANN baselines. Similar to Conversion/STDP results, end-to-end AGD training with spiking statistics (using surrogate gradient descent) for VGG5 and VGG3x2 of equivalent logical depth as VGG5 yields minimal accuracy loss ($\sim 2-3\%$) and large $EE$ gains ($\sim 15 \times$) in comparison to corresponding ANNs. However, for a $VGG7$ network with 7-layered depth, AGD fails to train an SNN end-to-end due to vanishing forward spike-propagation. Interestingly, a VGG3x4 network with similar logical depth of 7-layers as VGG7 and repeated computations not only trains well with AGD, but also yields higher accuracy than both VGG7/VGG3x4 ANN baselines. This implies that LIF neurons with spiking statistics have the potential of yielding more diversified computation profile with BackRes unrolling than ReLU neurons. In addition to accuracy and scalability benefits, SNNs with BackRes connections yield high $EE$ benefits as shown in Table \ref{table8} (due to the inherent \textit{`sparsifying'} effect) that point to their suitability for low-power hardware implementations.

\begin{table}
\caption{\textbf{Accuracy for AGD trained SNN of VGG3 topology (refer to last row in Table \ref{table10}) for different latency $T=5, 10, 25$ on CIFAR10 data.}}
\label{table10}
\centering
\begin{tabular}{|c|c|c|c|p{\textwidth}}
\hline
Model& $T=5$ & $T=10$ & $T=25$ \\
\hline
\makecell{VGG3\\(StochMax)} & 50.4 & 65.24 & 70.2\\
\hline
\makecell{VGG3\\(SoftMax)}  & 49.1& 64.44 & 67.1 \\
\hline
\makecell{VGG3\\(Topology)} & \multicolumn{3}{|c|}{\makecell{Input--Conv1(3,64,3x3/1)--Pool(2x2/2)\\--Conv2(64,64,3x3/1)--Pool(2x2/2)--\\FC1(4096,512)--FC2(512,10)}}\\
\hline
\end{tabular}
\end{table}

\begin{table}
\caption{\textbf{Accuracy and $EE$ benefits for AGD trained SNN with stochmax classifier on VGG5/VGG3x2 topology (refer to Table \ref{table9}) for different latency $T=25, 50$ on CIFAR10 data.}}
\label{table11}
\centering
\begin{tabular}{|c|c|c|c|c|c|p{\textwidth}}
\hline
Model& $T=25$ & $T=50$ & \makecell{$EE =$ \\ $\frac{E_{ANN}(1\times)}{E_{SNN}}$} &  \makecell{$EE =$ \\ $\frac{E_{SNN(softmax)}}{E_{SNN(stochmax)}}$}\\
&\multicolumn{2}{c|}{(-----Accuracy\%-----)}&\multicolumn{2}{|c|}{(----for $T=25$----)}\\
\hline
VGG5 & 75.26 & 75.92 & 23.83x & 1.62x\\
\hline
VGG3x2 & 72.62&73.17 & 31.88x & 1.97x\\
\hline
\end{tabular}
\end{table}

\begin{table}
\caption{\textbf{Accuracy and $EE$ benefits for STDP trained SNN with ConvNN classifier (Table \ref{table13}) appended to ResNet2, ResNet2x2, ResNet3 topology (refer to Table \ref{table7}) on CIFAR10 data. $EE_{Conv}$ considers the energy calculated only for the convolutional/pooling layers excluding the FC layers, $EE_{Full}$ considers the total energy of the network including the FC layers.}}
\label{table12}
\centering
\begin{tabular}{|c|c|c|c|c|c|p{\textwidth}}
\hline
Model& \makecell{ANN\\$T=1$} & \makecell{SNN\\$T=100$} & \makecell{$EE_{Conv}/EE_{Full} =$ \\ $\frac{E_{ANN}(1\times)}{E_{SNN}}$}\\
&\multicolumn{2}{c|}{(Accuracy\%)}&\\
\hline
ResNet2 & 83.5 & 77.92 & 1.64x/1.08x \\
\hline
ResNet3 & 79.85  & 76.52 & 1.81x/1.69x \\
\hline
ResNet2x2 & 83.2 & 80.1 & 10.56x/2.14x\\
\hline
\end{tabular}
\end{table}

\begin{table}
\setlength{\tabcolsep}{2pt}
\caption{\textbf{ConvNN classifier Network Topologies for STDP training methodology. Notations are same as that of Table \ref{table3}.}}
\label{table13}
\centering
\begin{tabular}{|c|c|c|p{\textwidth}}
\hline
Model & Configuration\\
\hline
\makecell{ConvNN\\ResNet2, ResNet2x2} & Input--Conv1(72,72,3x3/1)--Conv2(72,72,3x3/1)-- \\
&--Pool(2x2/2)--Conv3(72,144,3x3/1)--\\
&--Conv4(144,144,3x3/1)--Pool(2x2/2)--\\
&--FC1(2304,1024)--FC2(1024,10)\\
\hline
\makecell{ConvNN\\ResNet3} & Input--Conv1(108,108,3x3/1)--Conv2(108,108,3x3/1)--\\
&--Pool(2x2/2)--Conv3(108,216,3x3/1)--\\
&--Conv4(216,216,3x3/1)--Pool(2x2/2)--\\
&--FC1(3456,1024)--FC2(1024,10)\\
\hline
\end{tabular}
\end{table}

\subsection{Impact of Stochmax}
Stochmax is essentially a classification-performance improvement technique that can result in improved latency benefits. First, we show the impact of incorporating stochmax classifier for SNNs trained with AGD. Table \ref{table10} compares the accuracy of small VGG3 SNN trained with AGD for different latency $T$. Here, the $FC2$ layer of VGG3 topology is implemented as a softmax or stochmax classifier. We observe a consistent improvement in accuracy for stochmax implementations. In Table \ref{table11}, we show the accuracy results for SNNs of VGG5/VGG3x2 topology with stochmax classifiers. It is evident that stochmax improves the performance by $\sim3-4\%$ as compared to softmax implementations in Table \ref{table8}. In addition to accuracy, we also observe a larger gain in energy-efficiency with stochmax implementations. We find that conducting end-to-end AGD training with stochmax loss leads to sparser spiking activity across different layers of a network as compared to softmax. We believe this might be responsible for the efficiency gains. Further theoretical investigation is required to understand the role of loss optimization in a temporal processing landscape towards decreasing the spiking activity without affecting the gradient values. Table \ref{table10}, \ref{table11} results suggest stochmax as a viable technique for practical applications where we need to obtain higher accuracy and energy benefits with constrained latency or processing time.

Inclusion of stochmax classifier in SNNs trained with conversion/STDP training results in a slight improvement in accuracy $\sim 1-2\%$ for CIFAR10 data (for VGG7/ResNet3 topologies from Table \ref{table2}, \ref{table6}), respectively. Since stochmax is dissociated from the training process in both STDP/conversion, the latency and energy efficiency results are not affected. Note, all results shown in Table \ref{table2} - \ref{table8} use softmax classifier.

\subsection{Impact of Hybridization}

\begin{table}
\caption{\textbf{Accuracy, Total \# parameters and Energy Efficiency $EE$ for AGD trained SNN topologies  (refer Table \ref{table15}) with hybrid ReLU/LIF neurons of latency $T=25$ and corresponding ANN on CIFAR10 data.}}
\label{table14}
\centering
\begin{tabular}{|c|c|c|c|c|c|p{\textwidth}}
\hline
Model& \makecell{ANN\\$T=1$} & \makecell{SNN\\$T=25$} & \makecell{\#\\parameters} &\makecell{$EE =$ \\ $\frac{E_{ANN}(1\times)}{E_{SNN}}$}\\
&\multicolumn{2}{c|}{(Accuracy\%)}&&\\
\hline
VGG9 & 83.33 & 84.98 & 5.96M & 3.98x\\
\hline
VGG8x2 &  83.49 & 84.26 & 5.37M & 4.1x \\
\hline
\end{tabular}
\end{table}

\begin{table}
\setlength{\tabcolsep}{4pt}
\caption{\textbf{Network Topologies for AGD training methodology with hybrid layers and stochmax classifier at the end. Notations are same as that of Table \ref{table3}.}}
\label{table15}
\centering
\begin{tabular}{|c|c|c|p{\textwidth}}
\hline
Model & Configuration & BackRes\\
\hline
VGG9 & Input--Conv1(3,64,3x3/1)-ReLU-- & \multirow{4}{4em}{Not Applicable} \\
&--Conv2(64,64,3x3/1)-ReLU--Pool(2x2/2)--&\\
&--Conv3(64,128,3x3/1)-LIF--&\\
&--Conv4(128,128,3x3/1)-LIF--Pool(2x2/2)--&\\
&--Conv5(128,256,3x3/1)-LIF--&\\
&--Conv6(256, 256,3x3/1)-LIF--&\\
&--Conv7(256,256,3x3/1)-LIF--Pool(2x2/2)--&\\
&--FC1(4096,1024)-LIF--FC2(1024,10)&\\
\hline
VGG8x2 & Input--Conv1(3,64,3x3/1)-ReLU-- & \multirow{4}{4em}{\textcolor{red}{[Conv6]} repeated 2 times} \\
&--Conv2(64,64,3x3/1)-ReLU--Pool(2x2/2)-- &\\
&--Conv3(64,128,3x3/1)-LIF--&\\
&--Conv4(128,128,3x3/1)-LIF--Pool(2x2/2)--&\\
&--Conv5(128,256,3x3/1)-LIF--&\\
&--\textcolor{red}{Conv6(256, 256,3x3/1)}-LIF--&\\
&--\textcolor{red}{Conv6(256,256,3x3/1)}-LIF--Pool(2x2/2)--&\\
&--FC1(4096,1024)-LIF--FC2(1024,10)&\\
\hline
\end{tabular}
\end{table}
Except for Conversion, both STDP and AGD training techniques fail to yield high accuracy for deeper network implementations. While BackRes connections and Stochmax classifiers improve the accuracy, an SNN still lags behind its corresponding ANN in terms of performance. To improve the accuracy, we employ hybridization with partially ReLU and partially LIF neurons for SNN implementations. 

For STDP, we strengthen the classifier that is appended to the STDP trained convolutional layers to get better accuracy. Essentially, we replace the fully-connected layers $FC1, FC2$ of the topologies in Table \ref{table7} with a larger convolutional network $ConvNN$ ($ConvNN$ topology description is given in Table \ref{table13}). Table \ref{table12} shows the accuracy, $EE$ results for the STDP trained ResNet topologies appended now with corresponding $ConvNN$ and compared to a similar ANN baseline (say, ResNet2 ANN corresponds to an ANN with ResNet2 topology with FC layers replaced by ConvNN classifier from Table \ref{table13}). Strengthening the classifier hierarchy now results in higher accuracies ($\sim >75\%$) comparable to the ANN performance of Table \ref{table6}, while still lagging behind the ANN baseline of similar topology. However, the accuracy loss between ANN and SNN in this case reduces quite significantly ($>20\%$ loss in Table \ref{table6} to $\sim3\%$ loss in Table \ref{table12}). Similar to Table \ref{table6}, for $EE$, the gains considering only spiking layers are greater than that of the full network. 

For AGD, as discussed in Section VI, we hybridize our network with initial layers comprising of ReLU and latter layers of LIF neurons and perform end-to-end gradient descent. Table \ref{table14} shows the accuracy and $EE$ gain results for a VGG9, VGG8x2 model (topology description in Table \ref{table15}) with BackRes connection trained using hybridization for CIFAR10 dara. Note, only the first two convolutional layers $Conv1, Conv2$ use ReLU activation, while the remaining layers use LIF functionality. In addition, we use a stochmax classifier at the end instead of softmax to get better accuracy. Earlier, we saw that a 7-layered network could not be trained with AGD (see Table \ref{table8}). Inclusion of ReLU layers now allows a deep 9-layered network to be trained end-to-end while yielding considerable energy-efficiency gain with slightly improved accuracy ($\sim 1\%$ improvement in accuracy in SNN) in comparison to a corresponding ANN baseline (note, ANN baseline has ReLU activation in all layers). To have fair comparison between ANN and SNN, ANN baselines are trained without any batch normalziation or other regularization techniques. Including batch normalization and dropout in ANN training yields $\sim 86\%$ accuracy that is still fairly close to $\sim 85\%$ accuracy obtained with the SNN implementations. To calculate $EE$ gains in hybrid SNN implementations, we consider MAC energy for ReLU layers ($Conv1, Conv2$ in Table \ref{table14}) and AC energy for remaining LIF layers ($Conv3-Conv7(6)$ in Table \ref{table14}). VGG8x2 achieves equivalent logical depth as VGG9. Similar to earlier results, VGG8x2 yields slightly higher benefit than VGG9 on account of the \textit{`sparsifying' effect} induced by BackRes computations. 

Table \ref{table16} shows the results of a VGG13 model (topology description in Table \ref{table17})) trained with hybrid ReLU/LIF neuron layers on Imagenet dataset learn with end-to-end gradient descent. Interestingly, for Imagenet data, we had to use ReLU neuronal activations both in the beginning as well as at the end as shown in Table \ref{table17}. After some trial-and-error analysis, we found that training with more LIF neuronal layers for a complex dataset like Imagenet did not yield good performance. In case of a VGG13 network, converting the middle two layers into spiking LIF neurons yielded iso-accuracy as that of a fully-ReLU activation based ANN. Even with a minor portion of the network offering sparse neuronal spiking activity, we still observe $1.3 \times$ improvement in $EE$ with our hybrid model over the standard ANN. It is also worth mentioning that the spiking LIF neurons of the hybrid VGG13 network have a lower processing latency of $T=10$. We believe that using ReLU activations in majority of the VGG13 network enabled us to process the spiking layers at lower latency. We can expect higher $EE$ gains by adding suitable backward residual connections in the spiking layers to compensate for depth. It is evident that hybridization incurs a natural tradeoff between number of spiking/ReLU layers, processing latency, accuracy and energy-efficiency. Our analysis shows that hybridization can enable end-to-end backpropagation training for large-scale networks on complex datasets while yielding efficiency gains. Further investigation is required to evaluate the benefits of hybridization in large-scale setting by varying the tradeoff parameters.

\begin{table}
\caption{\textbf{Accuracy and Energy Efficiency $EE$ for AGD trained SNN topologies  (refer Table \ref{table17}) with hybrid ReLU/LIF neurons of latency $T=10$ and corresponding ANN on Imagenet data.}}
\label{table16}
\centering
\begin{tabular}{|c|c|c|c|p{\textwidth}}
\hline
Model& \makecell{ANN\\$T=1$} & \makecell{SNN\\$T=10$} & \makecell{$EE =$ \\ $\frac{E_{ANN}(1\times)}{E_{SNN}}$} \\
&\multicolumn{2}{c|}{(Accuracy\%)}&\\
\hline
VGG13 & \makecell{Top-1 69.9\\ Top-5 89.9} & \makecell{Top-1 67.6 \\ Top-5 88.23}  & 1.31x\\
\hline
\end{tabular}
\end{table}

\begin{table}
\setlength{\tabcolsep}{4pt}
\caption{\textbf{Network Topologies for AGD training methodology with hybrid layers and softmax classifier at the end for Imagenet Dataset. Notations are same as that of Table \ref{table3}.}}
\label{table17}
\centering
\begin{tabular}{|c|c|c|p{\textwidth}}
\hline
Model & Configuration & BackRes\\
\hline
VGG13 & Input--Conv1(3,64,3x3/1)-ReLU-- & \multirow{4}{4em}{Not Applicable} \\
&--Conv2(64,64,3x3/1)-ReLU--Pool(2x2/2)--&\\
&--Conv3(64,128,3x3/1)-ReLU--&\\
&--Conv4(128,128,3x3/1)-ReLU--Pool(2x2/2)--&\\
&--Conv5(128,256,3x3/1)-ReLU--&\\
&--Conv6(256,256,3x3/1)-ReLU--Pool(2x2/2)--&\\
&--Conv7(256, 512,3x3/1)-LIF--&\\
&--Conv8(512,512,3x3/1)-LIF--Pool(2x2/2)--&\\
&--Conv9(512,512,3x3/1)-ReLU--&\\
&--Conv10(512,512,3x3/1)-ReLU--Pool(2x2/2)--&\\
&--FC1(25088,4096)-ReLU--FC2(4096,4096)&\\
&--FC3(4096,1000)&\\
\hline
\end{tabular}
\end{table}

\section{Discussion \& Conclusion}
With the advent of Internet of Things (IoT) and the necessity to embed intelligence in devices that surround us (such, smart phones, health trackers), there is a need for novel computing solutions that offer energy benefits while yielding competitive performance. In this regard, SNNs driven by sparse event-driven processing hold promise for efficient hardware implementation of real-world applications. However, training SNNs for large-scale tasks still remains a challenge. In this work, we outlined the limitation of the three widely used SNN training methodologies (Conversion, AGD training and STDP), in terms of, \textit{scalability, latency and accuracy}, and proposed novel solutions to overcome them. 

We propose using backward residual (or BackRes) connections to achieve logically deep SNNs with shared network computations and features that can approach the accuracy of fully-deep SNNs. We show that all three training methods benefit from the BackRes connection inclusion in the network configuration, especially, gaining in terms of energy-efficiency ($\sim 10 \times - 100 \times$) while yielding iso-accuracy with that of an ANN of similar configuration. We also find that BackRes connections induce a \textit{sparsifying effect} on overall network activity of an SNN, thereby, expending lower energy ($\sim 1.8-3.5 \times$ lower) than an equivalent depth full-layered SNN. In summary, BackRes connections address the scalability limitations of an SNN that arise due to depth incompatibility and \textit{vanishing spike-propagation} of different training techniques.

We propose using stochastic softmax (or stochmax) to improve the prediction capability of an SNN, specifically, for AGD training method that uses end-to-end spike-based backpropagation. We find a significant improvement in accuracy ($\sim 2-3\%$) with stochmax inclusion even for lower latency or processing time period. Further, stochmax loss based backpropagation results in lower spiking activity than the conventional softmax loss. Combining the advantages of lower latency and sparser activity, we get higher energy-efficiency improvements ($\sim 1.6-2 \times$) with stochmax SNNs as compared to softmax SNNs. Conversion/STDP training do not benefit in terms of efficiency and latency from stochmax inclusion since the training in these cases are performed fully/partially with ANN computations.

The third technique we propose is using a hybrid architecture with partly-ReLU-and-partly-LIF computations in order to improve the accuracy obtained with STDP/AGD training methods. We find that hybridization leads to improved accuracy at lower latency for AGD/STDP methods, even circumventing the inadequacy of training very deep networks. The accuracies observed for CIFAR10 ($\sim 80\%/85\%$) with STDP/AGD on hybrid SNN architectures are in fact comparable/better than ANNs of similar configuration. We would like to note that hybridization also offers significant energy-efficiency improvement ($\sim 4 \times$) over a fully ReLU-based ANN. In fact, using hybridization, we trained a deep VGG13 model on Imagenet data and obtained iso-accuracy as that of its ANN counterpart with reasonable energy-efficiency gains. There are interesting possibilities of performing distributed edge-cloud intelligence with such hybrid SNN-ANN architecture where, SNN layers can be implemented on resource-constrained edge devices and ANN layers on the cloud. 

Finally, SNNs are a prime candidate today towards enabling low-powererd ubiquitous intelligence. In this paper, we show the benefit of using good practices while configuring spiking networks to overcome their inherent training limitations, while, gaining in terms of energy-efficiency, latency and accuracy for image recognition applications. In the future, we will investigate the extension of the proposed methods for training recurrent models for natural language or video processing tasks. Further, conducting reinforcement learning with the above proposed techniques to analyze the advantages that SNNs offer is another possible future work direction.


%

%
%
%
%
\section*{Acknowledgment}
This work was supported in part by C-BRIC, Center for Brain-inspired Computing, a JUMP center sponsored by DARPA and SRC, by the Semiconductor Research Corporation, the National Science Foundation, Intel Corporation, the Vannevar Bush Faculty Fellowship and the U.K. Ministry of Defense under Agreement Number W911NF-16-3-0001.
%
%
%
%



\bibliographystyle{IEEEtran}
\bibliography{IEEEabrv,refs}

\begin{thebibliography}{10}
\providecommand{\url}[1]{#1}
\csname url@samestyle\endcsname
\providecommand{\newblock}{\relax}
\providecommand{\bibinfo}[2]{#2}
\providecommand{\BIBentrySTDinterwordspacing}{\spaceskip=0pt\relax}
\providecommand{\BIBentryALTinterwordstretchfactor}{4}
\providecommand{\BIBentryALTinterwordspacing}{\spaceskip=\fontdimen2\font plus
\BIBentryALTinterwordstretchfactor\fontdimen3\font minus
  \fontdimen4\font\relax}
\providecommand{\BIBforeignlanguage}[2]{{%
\expandafter\ifx\csname l@#1\endcsname\relax
\typeout{** WARNING: IEEEtran.bst: No hyphenation pattern has been}%
\typeout{** loaded for the language `#1'. Using the pattern for}%
\typeout{** the default language instead.}%
\else
\language=\csname l@#1\endcsname
\fi
#2}}
\providecommand{\BIBdecl}{\relax}
\BIBdecl

\bibitem{indiveri2011frontiers}
G.~Indiveri and T.~K. Horiuchi, ``Frontiers in neuromorphic engineering,''
  \emph{Frontiers in neuroscience}, vol.~5, p. 118, 2011.

\bibitem{pfeiffer2018deep}
M.~Pfeiffer and T.~Pfeil, ``Deep learning with spiking neurons: opportunities
  and challenges,'' \emph{Frontiers in neuroscience}, vol.~12, 2018.

\bibitem{cao2015spiking}
Y.~Cao, Y.~Chen, and D.~Khosla, ``Spiking deep convolutional neural networks
  for energy-efficient object recognition,'' \emph{International Journal of
  Computer Vision}, vol. 113, no.~1, pp. 54--66, 2015.

\bibitem{panda2016unsupervised}
P.~Panda and K.~Roy, ``Unsupervised regenerative learning of hierarchical
  features in spiking deep networks for object recognition,'' in \emph{2016
  International Joint Conference on Neural Networks (IJCNN)}.\hskip 1em plus
  0.5em minus 0.4em\relax IEEE, 2016, pp. 299--306.

\bibitem{sengupta2016hybrid}
A.~Sengupta, A.~Banerjee, and K.~Roy, ``Hybrid spintronic-cmos spiking neural
  network with on-chip learning: Devices, circuits, and systems,''
  \emph{Physical Review Applied}, vol.~6, no.~6, p. 064003, 2016.

\bibitem{ankit2017resparc}
A.~Ankit, A.~Sengupta, P.~Panda, and K.~Roy, ``Resparc: A reconfigurable and
  energy-efficient architecture with memristive crossbars for deep spiking
  neural networks,'' in \emph{Proceedings of the 54th Annual Design Automation
  Conference 2017}.\hskip 1em plus 0.5em minus 0.4em\relax ACM, 2017, p.~27.

\bibitem{indiveri2015neuromorphic}
G.~Indiveri, F.~Corradi, and N.~Qiao, ``Neuromorphic architectures for spiking
  deep neural networks,'' in \emph{2015 IEEE International Electron Devices
  Meeting (IEDM)}.\hskip 1em plus 0.5em minus 0.4em\relax IEEE, 2015, pp. 4--2.

\bibitem{diehl2015fast}
P.~U. Diehl, D.~Neil, J.~Binas, M.~Cook, S.-C. Liu, and M.~Pfeiffer,
  ``Fast-classifying, high-accuracy spiking deep networks through weight and
  threshold balancing,'' in \emph{2015 International Joint Conference on Neural
  Networks (IJCNN)}.\hskip 1em plus 0.5em minus 0.4em\relax IEEE, 2015, pp.
  1--8.

\bibitem{lee2016training}
J.~H. Lee, T.~Delbruck, and M.~Pfeiffer, ``Training deep spiking neural
  networks using backpropagation,'' \emph{Frontiers in neuroscience}, vol.~10,
  p. 508, 2016.

\bibitem{o2013real}
P.~O'Connor, D.~Neil, S.-C. Liu, T.~Delbruck, and M.~Pfeiffer, ``Real-time
  classification and sensor fusion with a spiking deep belief network,''
  \emph{Frontiers in neuroscience}, vol.~7, p. 178, 2013.

\bibitem{kheradpisheh2018stdp}
S.~R. Kheradpisheh, M.~Ganjtabesh, S.~J. Thorpe, and T.~Masquelier,
  ``Stdp-based spiking deep convolutional neural networks for object
  recognition,'' \emph{Neural Networks}, vol.~99, pp. 56--67, 2018.

\bibitem{masquelier2009competitive}
T.~Masquelier, R.~Guyonneau, and S.~J. Thorpe, ``Competitive stdp-based spike
  pattern learning,'' \emph{Neural computation}, vol.~21, no.~5, pp.
  1259--1276, 2009.

\bibitem{lee2018training}
C.~Lee, P.~Panda, G.~Srinivasan, and K.~Roy, ``Training deep spiking
  convolutional neural networks with stdp-based unsupervised pre-training
  followed by supervised fine-tuning,'' \emph{Frontiers in neuroscience},
  vol.~12, 2018.

\bibitem{lee2018deep}
C.~Lee, G.~Srinivasan, P.~Panda, and K.~Roy, ``Deep spiking convolutional
  neural network trained with unsupervised spike timing dependent plasticity,''
  \emph{IEEE Transactions on Cognitive and Developmental Systems}, 2018.

\bibitem{srinivasan2018stdp}
G.~Srinivasan, P.~Panda, and K.~Roy, ``Stdp-based unsupervised feature learning
  using convolution-over-time in spiking neural networks for energy-efficient
  neuromorphic computing,'' \emph{ACM Journal on Emerging Technologies in
  Computing Systems (JETC)}, vol.~14, no.~4, p.~44, 2018.

\bibitem{panda2017asp}
P.~Panda, J.~M. Allred, S.~Ramanathan, and K.~Roy, ``Asp: Learning to forget
  with adaptive synaptic plasticity in spiking neural networks,'' \emph{IEEE
  Journal on Emerging and Selected Topics in Circuits and Systems}, vol.~8,
  no.~1, pp. 51--64, 2017.

\bibitem{diehl2015unsupervised}
P.~U. Diehl and M.~Cook, ``Unsupervised learning of digit recognition using
  spike-timing-dependent plasticity,'' \emph{Frontiers in computational
  neuroscience}, vol.~9, p.~99, 2015.

\bibitem{masquelier2007unsupervised}
T.~Masquelier and S.~J. Thorpe, ``Unsupervised learning of visual features
  through spike timing dependent plasticity,'' \emph{PLoS computational
  biology}, vol.~3, no.~2, p. e31, 2007.

\bibitem{hunsberger2015spiking}
E.~Hunsberger and C.~Eliasmith, ``Spiking deep networks with lif neurons,''
  \emph{arXiv preprint arXiv:1510.08829}, 2015.

\bibitem{bellec2018long}
G.~Bellec, D.~Salaj, A.~Subramoney, R.~Legenstein, and W.~Maass, ``Long
  short-term memory and learning-to-learn in networks of spiking neurons,'' in
  \emph{Advances in Neural Information Processing Systems}, 2018, pp. 787--797.

\bibitem{neftci2019surrogate}
E.~O. Neftci, H.~Mostafa, and F.~Zenke, ``Surrogate gradient learning in
  spiking neural networks,'' \emph{arXiv preprint arXiv:1901.09948}, 2019.

\bibitem{mostafa2017supervised}
H.~Mostafa, ``Supervised learning based on temporal coding in spiking neural
  networks,'' \emph{IEEE transactions on neural networks and learning systems},
  vol.~29, no.~7, pp. 3227--3235, 2017.

\bibitem{sengupta2019going}
A.~Sengupta, Y.~Ye, R.~Wang, C.~Liu, and K.~Roy, ``Going deeper in spiking
  neural networks: Vgg and residual architectures,'' \emph{Frontiers in
  neuroscience}, vol.~13, 2019.

\bibitem{severa2019training}
W.~Severa, C.~M. Vineyard, R.~Dellana, S.~J. Verzi, and J.~B. Aimone,
  ``Training deep neural networks for binary communication with the whetstone
  method,'' \emph{Nature Machine Intelligence}, vol.~1, no.~2, p.~86, 2019.

\bibitem{srinivasan2019restocnet}
G.~Srinivasan and K.~Roy, ``Restocnet: Residual stochastic binary convolutional
  spiking neural network for memory-efficient neuromorphic computing,''
  \emph{Frontiers in Neuroscience}, vol.~13, p. 189, 2019.

\bibitem{deng2009imagenet}
J.~Deng, W.~Dong, R.~Socher, L.-J. Li, K.~Li, and L.~Fei-Fei, ``Imagenet: A
  large-scale hierarchical image database,'' in \emph{2009 IEEE conference on
  computer vision and pattern recognition}.\hskip 1em plus 0.5em minus
  0.4em\relax Ieee, 2009, pp. 248--255.

\bibitem{krizhevsky2010convolutional}
A.~Krizhevsky and G.~Hinton, ``Convolutional deep belief networks on
  cifar-10,'' \emph{Unpublished manuscript}, vol.~40, no.~7, pp. 1--9, 2010.

\bibitem{lecun2010mnist}
Y.~LeCun, C.~Cortes, and C.~Burges, ``Mnist handwritten digit database,''
  \emph{AT\&T Labs [Online]. Available: http://yann. lecun. com/exdb/mnist},
  vol.~2, p.~18, 2010.

\bibitem{simonyan2014very}
K.~Simonyan and A.~Zisserman, ``Very deep convolutional networks for
  large-scale image recognition,'' \emph{arXiv preprint arXiv:1409.1556}, 2014.

\bibitem{he2016deep}
K.~He, X.~Zhang, S.~Ren, and J.~Sun, ``Deep residual learning for image
  recognition,'' in \emph{Proceedings of the IEEE conference on computer vision
  and pattern recognition}, 2016, pp. 770--778.

\bibitem{lee2019enabling}
C.~Lee, S.~S. Sarwar, and K.~Roy, ``Enabling spike-based backpropagation in
  state-of-the-art deep neural network architectures,'' \emph{arXiv preprint
  arXiv:1903.06379}, 2019.

\bibitem{sengupta2017encoding}
A.~Sengupta and K.~Roy, ``Encoding neural and synaptic functionalities in
  electron spin: A pathway to efficient neuromorphic computing,'' \emph{Applied
  Physics Reviews}, vol.~4, no.~4, p. 041105, 2017.

\bibitem{wang2017memristors}
Z.~Wang, S.~Joshi, S.~E. Savel’ev, H.~Jiang, R.~Midya, P.~Lin, M.~Hu, N.~Ge,
  J.~P. Strachan, Z.~Li \emph{et~al.}, ``Memristors with diffusive dynamics as
  synaptic emulators for neuromorphic computing,'' \emph{Nature materials},
  vol.~16, no.~1, p. 101, 2017.

\bibitem{van2017non}
Y.~van~de Burgt, E.~Lubberman, E.~J. Fuller, S.~T. Keene, G.~C. Faria,
  S.~Agarwal, M.~J. Marinella, A.~A. Talin, and A.~Salleo, ``A non-volatile
  organic electrochemical device as a low-voltage artificial synapse for
  neuromorphic computing,'' \emph{Nature materials}, vol.~16, no.~4, p. 414,
  2017.

\bibitem{perez2010neuromorphic}
J.~A. P{\'e}rez-Carrasco, C.~Zamarre{\~n}o-Ramos, T.~Serrano-Gotarredona, and
  B.~Linares-Barranco, ``On neuromorphic spiking architectures for asynchronous
  stdp memristive systems,'' in \emph{Proceedings of 2010 IEEE International
  Symposium on Circuits and Systems}.\hskip 1em plus 0.5em minus 0.4em\relax
  IEEE, 2010, pp. 1659--1662.

\bibitem{linares2011spike}
B.~Linares-Barranco, T.~Serrano-Gotarredona, L.~A. Camu{\~n}as-Mesa, J.~A.
  Perez-Carrasco, C.~Zamarre{\~n}o-Ramos, and T.~Masquelier, ``On
  spike-timing-dependent-plasticity, memristive devices, and building a
  self-learning visual cortex,'' \emph{Frontiers in neuroscience}, vol.~5,
  p.~26, 2011.

\bibitem{lecun2015deep}
Y.~LeCun, Y.~Bengio, and G.~Hinton, ``Deep learning,'' \emph{nature}, vol. 521,
  no. 7553, p. 436, 2015.

\bibitem{szegedy2015going}
C.~Szegedy, W.~Liu, Y.~Jia, P.~Sermanet, S.~Reed, D.~Anguelov, D.~Erhan,
  V.~Vanhoucke, and A.~Rabinovich, ``Going deeper with convolutions,'' in
  \emph{Proceedings of the IEEE conference on computer vision and pattern
  recognition}, 2015, pp. 1--9.

\bibitem{kubilius2018cornet}
J.~Kubilius, M.~Schrimpf, A.~Nayebi, D.~Bear, D.~L. Yamins, and J.~J. DiCarlo,
  ``Cornet: modeling the neural mechanisms of core object recognition,''
  \emph{BioRxiv}, p. 408385, 2018.

\bibitem{lee2018dropmax}
H.~B. Lee, J.~Lee, S.~Kim, E.~Yang, and S.~J. Hwang, ``Dropmax: Adaptive
  variational softmax,'' in \emph{Advances in Neural Information Processing
  Systems}, 2018, pp. 919--929.

\bibitem{abadi2016tensorflow}
M.~Abadi, P.~Barham, J.~Chen, Z.~Chen, A.~Davis, J.~Dean, M.~Devin,
  S.~Ghemawat, G.~Irving, M.~Isard \emph{et~al.}, ``Tensorflow: A system for
  large-scale machine learning,'' in \emph{12th $\{$USENIX$\}$ Symposium on
  Operating Systems Design and Implementation ($\{$OSDI$\}$ 16)}, 2016, pp.
  265--283.

\bibitem{werbos1990backpropagation}
P.~J. Werbos \emph{et~al.}, ``Backpropagation through time: what it does and
  how to do it,'' \emph{Proceedings of the IEEE}, vol.~78, no.~10, pp.
  1550--1560, 1990.

\bibitem{han2015learning}
S.~Han, J.~Pool, J.~Tran, and W.~Dally, ``Learning both weights and connections
  for efficient neural network,'' in \emph{Advances in neural information
  processing systems}, 2015, pp. 1135--1143.

\bibitem{han2015deep}
S.~Han, H.~Mao, and W.~J. Dally, ``Deep compression: Compressing deep neural
  networks with pruning, trained quantization and huffman coding,'' \emph{arXiv
  preprint arXiv:1510.00149}, 2015.

\end{thebibliography}
\end{document}